\documentclass[11pt, letterpaper, logo]{googledeepmind}

\PassOptionsToPackage{comma,numbers,sort,compress}{natbib}

\usepackage[comma,numbers,sort,compress]{natbib}

\usepackage{hyperref}[citecolor=magenta]

\hypersetup{
    colorlinks = true,
    citecolor = {magenta},
}

\pdftrailerid{redacted}

\makeatletter
\renewcommand\bibentry[1]{\nocite{#1}{\frenchspacing\@nameuse{BR@r@#1\@extra@b@citeb}}}
\makeatother

\usepackage{kantlipsum, lipsum}
\usepackage{dsfont}
\usepackage{gdm-colors}
\usepackage{subfigure}
\usepackage{bbm}
\usepackage{wrapfig}
\usepackage{colortbl}
\usepackage{xcolor}
\usepackage{graphicx}
\usepackage{caption}
\usepackage[percent]{overpic}

\usepackage[most,skins,theorems]{tcolorbox}

\tcbset{
  aibox/.style={
    width=\linewidth,
    top=8pt,
    bottom=4pt,
    colback=blue!6!white,
    colframe=black,
    colbacktitle=black,
    enhanced,
    center,
    attach boxed title to top left={yshift=-0.1in,xshift=0.15in},
    boxed title style={boxrule=0pt,colframe=white,},
  }
}
\newtcolorbox{AIbox}[2][]{aibox,title=#2,#1}
\definecolor{lightblue}{rgb}{0.22,0.45,0.70}

\graphicspath{{figures/}}

\title{Inference-Time Scaling for Diffusion Models beyond Scaling Denoising Steps}

\correspondingauthor{nm3607@nyu.edu, sainx@google.com}

\author[$\vardiamondsuit$, $\clubsuit$, 1]{Nanye Ma}
\author[$\vardiamondsuit$, $\clubsuit$, 2]{Shangyuan Tong}
\author[3]{Haolin Jia}
\author[3]{Hexiang Hu}
\author[3]{Yu-Chuan Su}
\author[3]{Mingda Zhang}
\author[3]{Xuan Yang}
\author[3]{Yandong Li}
\author[2]{Tommi Jaakkola}
\author[3]{Xuhui Jia}
\author[1,3]{Saining Xie}

\affil[$\vardiamondsuit$]{Equal contribution}
\affil[1]{NYU}
\affil[2]{MIT}
\affil[3]{Google}
\affil[$\clubsuit$]{Work done during an internship at Google}
\vspace{-0.5cm}

\begin{abstract}
\vspace{-0.4cm}
Generative models have made significant impacts across various domains, largely due to their ability to scale during training by increasing data, computational resources, and model size, a phenomenon characterized by the scaling laws. Recent research has begun to explore inference-time scaling behavior in Large Language Models (LLMs), revealing how performance can further improve with additional computation during inference. 
Unlike LLMs, diffusion models inherently possess the flexibility to adjust inference-time computation via the number of denoising steps, although the performance gains typically flatten after a few dozen.
In this work, we explore the inference-time scaling behavior of diffusion models \textit{beyond} increasing denoising steps and investigate how the generation performance can further improve with increased computation.
Specifically, we consider a search problem aimed at identifying better noises for the diffusion sampling process. We structure the design space along two axes: the verifiers used to provide feedback, and the algorithms used to find better noise candidates. Through extensive experiments on class-conditioned and text-conditioned image generation benchmarks, our findings reveal that increasing inference-time compute leads to substantial improvements in the quality of samples generated by diffusion models, and with the complicated nature of images, combinations of the components in the framework can be specifically chosen to conform with different application scenario.
\end{abstract}

\begin{document}

\maketitle

\newcommand{\expect}[2]{\mathds{E}_{{#1}} \left[ {#2} \right]}
\newcommand{\myvec}[1]{\boldsymbol{#1}}
\newcommand{\myvecsym}[1]{\boldsymbol{#1}}
\newcommand{\vx}{\myvec{x}}
\newcommand{\vy}{\myvec{y}}
\newcommand{\vz}{\myvec{z}}
\newcommand{\vtheta}{\myvecsym{\theta}}

\definecolor{mygreen}{RGB}{144,175,115}
\definecolor{myyellow}{RGB}{234,172,84}
\definecolor{myred}{RGB}{227,90,80}
\definecolor{myblue}{RGB}{103,139,245}

\vspace{-0.25cm}
\section{Introduction}
\vspace{-0.25cm}
\label{sec:intro}

\begin{figure}[t]
    \centering
    \includegraphics[width=0.75\linewidth]{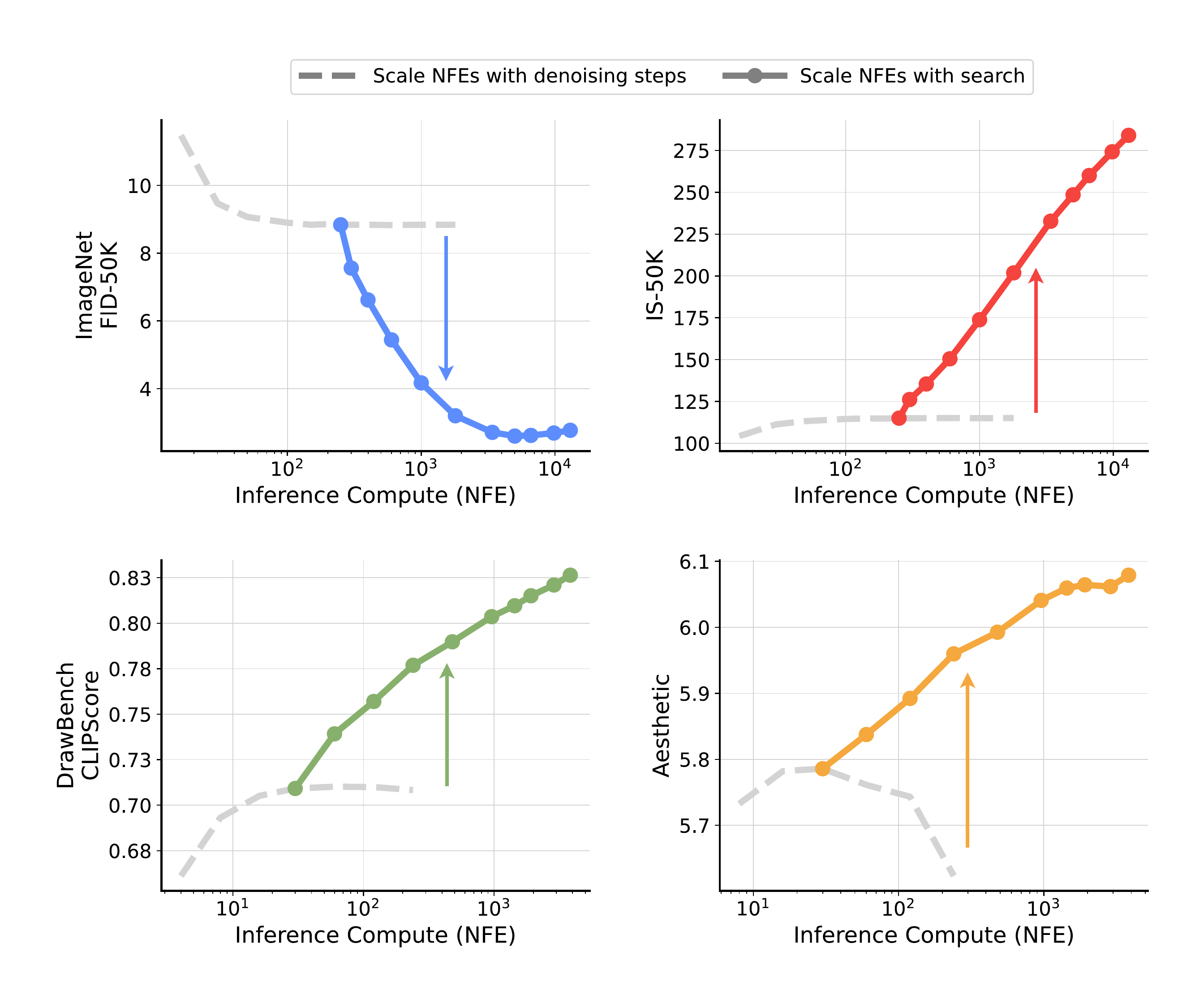}

    \vspace{-0.8cm}
    \caption{\footnotesize{\textbf{\emph{Inference scaling beyond increasing denoising steps.}} We demonstrate the performance with respect to FID \textcolor[RGB]{103,139,245}{$\boldsymbol{\downarrow}$}, IS \textcolor{myred}{$\boldsymbol{\uparrow}$} on ImageNet, and CLIPScore \textcolor{mygreen}{$\boldsymbol{\uparrow}$}, Aesthetic Score \textcolor{myyellow}{$\boldsymbol{\uparrow}$} on DrawBench. Our search framework exhibits substantial improvements in all settings over purely scaling NFEs with increasing denoising steps.}}
    \vspace{-0.5cm}
    \label{fig:teaser}
\end{figure}

Generative models have transformed various fields, including language~\citep{achiam2023gpt, team2023gemini, touvron2023llama}, vision~\citep{rombach2022high, ramesh2021zero}, and biology~\citep{watson2023novo}, by learning to sample from the underlying data distributions. A key factor in their success is their ability to scale up during training by increasing data volumes, computational resources, and model sizes. This training-time scaling behavior, often described as \textit{Scaling Laws}~\citep{kaplan2020scaling, hoffmann2022training}, predicts how performance improves as the models grow larger, consume more data, and are trained for longer time, guiding the development of increasingly capable generative models.

Recently, in Large Language Models (LLMs), the study on scaling has expanded to inference-time~\citep{snell2024scaling, brown2024large, wu2024empirical}. By allocating more compute during inference, often through sophisticated search processes, these works show that LLMs can produce higher-quality and more contextually appropriate responses~\citep{wei2022chain, wang2022self, yao2024tree, gandhi2024stream, su2024dualformer}. Inference-time scaling opens new avenues for improving model performance when additional resources become available after training.

Diffusion models~\citep{sohl2015deep, ho2020denoising, song2021scorebased}, trained to remove noises from data, are a class of generative models that dominates the continuous data domains such as images~\citep{esser2024scaling}, audio~\citep{schneider2023archisound}, and videos~\citep{polyak2024movie}. 
To generate a single sample, their generation process usually starts from pure noise and requires multiple forward passes of trained models to denoise and obtain clean data. These forward passes are thus dubbed \textit{denoising steps}.
Since the number of denoising steps can be adjusted to trade sample quality for computational cost, the generation process of diffusion models naturally provides flexibility in allocating inference-time computation budget. Under the context of generative models, such computation budget is also commonly measured by the \textit{number of function evaluations} (NFE), to ensure a reasonable comparison with other families of models that use iterative sampling processes but without denoising capabilities~\citep{tian2024var, yu2024magvit},

Empirical observations~\citep{song2021scorebased, karras2022elucidating, song2021denoising} have indicated that by investing compute into denoising steps alone,  performance gains tend to plateau after a certain NFEs, limiting the benefits of scaling up computation during inference.  Therefore, previous work on diffusion models has long focused on maintaining high performance while making NFEs as small as possible for efficiency during inference time~\citep{salimans2022progressive, song2023consistency}. We, on the other hand, are interested in the opposite frontier.

Compared to LLMs, diffusion models work with explicit randomness that comes from the noises injected either as initial samples or during the sampling process~\citep{song2021scorebased, xu2023restart}. It has been shown that these noises are not created equal~\citep{ahn2024noiseworthdiffusionguidance, qi2024not}, i.e., some lead to better generations than others. This observation extends an additional dimension to scale NFEs other than increasing denoising steps - searching for better noises in sampling. Rather than solely allocating NFEs for denoising steps, which often leads to a quick performance plateau, this work investigates methods to effectively utilize compute during inference through search, thereby improving the performance and scalability of diffusion models at inference time (Figure~\ref{fig:teaser}). We primarily consider two design axes in our search framework: \emph{verifiers} used to provide feedback in search, and \emph{algorithms} used to find better noise candidates, following terminologies used in LLMs~\citep{snell2024scaling}.

For \textit{verifiers}, we consider the three different settings, which are meant to simulate three different use cases: (1) where we have privileged information about how the final evaluation is carried out; (2) where we have conditioning information for guiding the generation; (3) where we have no extra information available. For \textit{algorithms}, we consider (1) Random Search, which simply selects the best from a fixed set of candidates; (2) Zero-Order Search, which leverages verifiers feedback to iteratively refine noise candidates; (3) Search over Paths, which leverages verifiers feedback to iteratively refine diffusion sampling trajectories.

We first walk over these design choices in the relatively simple setting of class-conditioned generation on ImageNet and demonstrate their effectiveness, providing an instantiation of our framework. Then we carry these design choices over to the larger-scale text-conditioned generation setting and evaluate our proposed framework. Due to the complex nature of images and the rich information text conditionings contain, more holistic evaluations of generation quality are required~\citep{lee2024holistic}. We therefore employ multiple verifiers for scaling inference-time compute in search. This also enables us to probe into the ``biases'' each verifier possesses, and how well they are aligned with the generation tasks. To alleviate overfitting to a single verifier, we also experiment with an ensemble of verifiers and showcase its good generalizability across different benchmarks.

Our contributions are summarized as follows:
\begin{itemize}
    \item We propose a fundamental framework for inference-time scaling of diffusion models. We show that scaling NFEs through search can lead to substantial improvement across generation tasks and model sizes beyond increasing denoising steps. In addition, we conduct a comprehensive empirical analysis of how inference-time compute budgets affect scaling performance.
    \item We identify two key design axes in the proposed search framework: verifiers, which provide feedback, and algorithms, which find better noise candidates. We examine how different verifier–algorithm combinations perform across various tasks, and our findings indicate that no single configuration is universally optimal; each task instead necessitates a distinct search setup to achieve the best scaling performance.
    \item We conduct extensive analysis on the alignment between verifiers and different generation tasks. Our results shed light on the biases embedded inside different verifiers and the necessity for a specifically designed verifier in each distinct vision generation task.
\end{itemize}

\vspace{-0.2cm}
\section{Background and Motivation}
\label{sec:background}
\vspace{-0.2cm}

\textbf{Diffusion models.} Diffusion models~\citep{sohl2015deep, ho2020denoising, song2021scorebased} and more generally flow-based models~\citep{albergo2023building, liu2023flow, lipman2023flow} are a family of generative models that learn to reverse a reference ``noising'' process. We follow the notations presented in EDM~\citep{karras2022elucidating}, and let the data distribution we want to model be $p_{\text{data}}(\myvec{x})$ with standard deviation $\sigma_{\text{data}}$. We consider a reference process that injects different levels of i.i.d. Gaussian noise to the clean data, specified by its standard deviation $\sigma$, and we denote these mollified distributions as $p(\myvec{x}; \sigma)$. The terminal noise level $\sigma_{\text{max}}\gg\sigma_{\text{data}}$$^1$
\fancypagestyle{footnote1}{\fancyfoot[L]{\footerfont \itshape $^1$Flow-based models effectively have 
$\sigma_{\text{max}}=\infty$.}}
\thispagestyle{footnote1}
lets this reference process destroy practically all information of the data $p(\myvec{x}; \sigma_{\text{max}})\approx \mathcal{N}(\myvec{0}, \sigma_{\text{max}}^2\myvec{I})$. Generation then starts from a pure noise, and simulates some differential equation to progressively denoise the sample to a clean one. Specifically for diffusion models, the underlying vector field is closely related to the score functions $\nabla_{\myvec{x}}\log p(\myvec{x};\sigma)$ at different noise levels. Often an ordinary differential equation (ODE)~\citep{song2021scorebased} or a stochastic differential equation (SDE)~\citep{anderson1982reverse} is used during the sampling process:
\begin{align*}
    &\text{ODE} &&\mathrm{d}\myvec{x} = -\dot{\sigma}(t)\sigma(t)\nabla_{\myvec{x}}\log p(\myvec{x};\sigma(t))\mathrm{d}t \\
    &\text{SDE} &&\mathrm{d}\myvec{x} = -2\dot{\sigma}(t)\sigma(t)\nabla_{\myvec{x}}\log p(\myvec{x};\sigma(t))\mathrm{d}t + \sqrt{2\dot{\sigma}(t)\sigma(t)} \mathrm{d}W_t
\end{align*}
where $\sigma(t)$ is the $\sigma$ schedule w.r.t. time, and $W_t$ is the standard Wiener process. Diffusion models $D_\theta$ are trained to approximate the ground truth score functions.

\textbf{Innate scaling at inference time.} One remarkable property of diffusion models is their innate flexibility to allocate varied compute at inference time for the same task. Because they are trained to approximate the underlying vector field, diffusion models are evaluated multiple times at different noise levels for a single generation. Effectively, the sampling process can be understood as a rolled-out, much larger model, that is stably trained only parts at a time. This mismatch in capacity between training and inference time is one of the key characteristics that separate diffusion models and other generative models like GANs~\citep{goodfellow2014generative} and VAEs~\citep{kingma2013auto}. Investing more compute to denoising steps generally leads to better generations, but with diminishing benefits, due to the accumulation of both approximation and discretization errors~\citep{xu2023restart}. Therefore, for diffusion models to scale more at inference time, a new framework needs to be designed.

\textbf{Randomness from noise.} In theory, there is explicit randomness in the sampling of diffusion models: the randomly drawn initial noise, and the optional subsequent noise injected via procedures like SDE~\citep{song2021scorebased} and Restart Sampling~\citep{xu2023restart}. Nonetheless, because the model evaluations are inherently deterministic, there is a fixed mapping from these noises to the final samples$^2$. It has been shown that some noises are better than others~\citep{qi2024not,ahn2024noiseworthdiffusionguidance}, suggesting that it is possible to push the inference time scaling limit by devoting more NFEs to finding the more preferable noises for sampling.

\vspace{-0.25cm}
\section{How to Scale at Inference Time}
\label{sec:imagenet}
\vspace{-0.25cm}

\begin{figure}[t!]
    \centering
    \includegraphics[width=\linewidth]{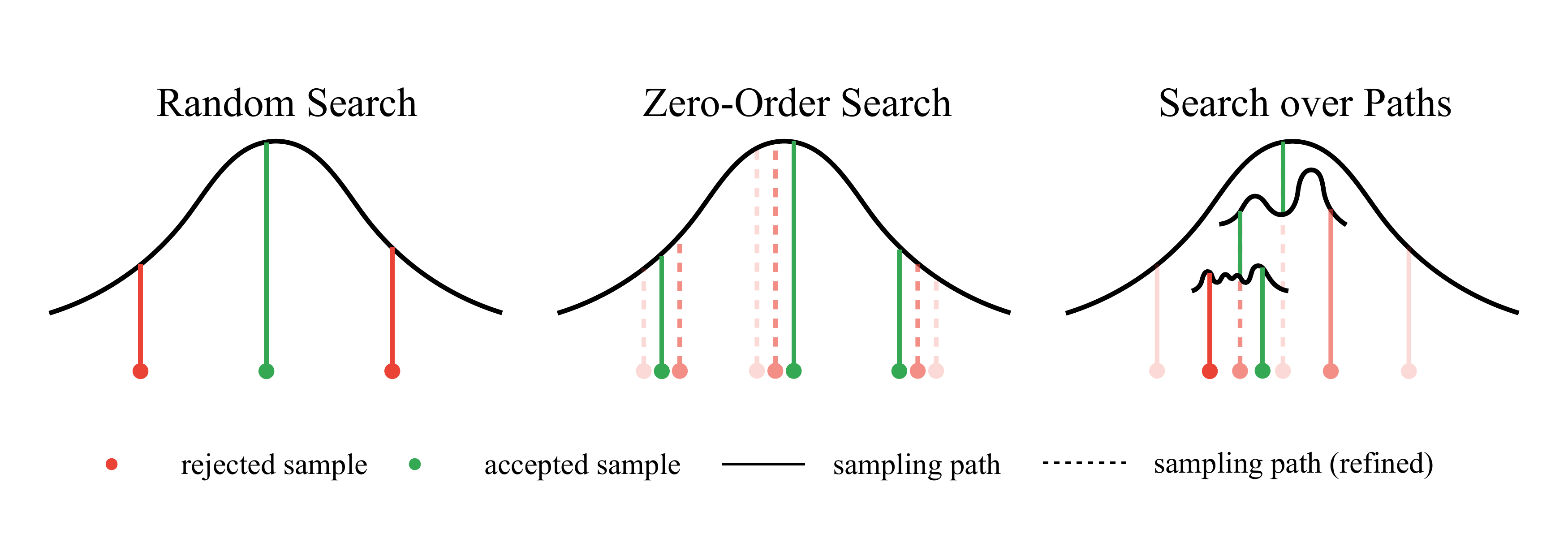}
    \vspace{-1.5cm}
    \caption{\footnotesize{\textbf{\emph{Illustration of Search Algorithms.}}} \textbf{Left:} \textit{Random Search} selects the best sample according to the verifier score and rejects the rest. \textbf{Center:} \textit{Zero-Order Search} samples $N$ candidates in the neighborhood of the pivot noise at each step, and selects the best one according to the verifier to continue the search from. \textbf{Right:} \textit{Search over Paths} sample noises at intermediate sampling steps to add to current samples to expand the sampling trajectories, and select the best one to continue the search.}
    \label{fig:overview}
    \vspace{-0.5cm}
\end{figure}

\fancypagestyle{footnote2}{\fancyfoot[L]{\footerfont \itshape $^2$Technically we also need to fix the same NFEs in denoising steps, but in practice this requirement is often quite loose (see Section~\ref{sec:dimension} for detail).}}
\thispagestyle{footnote2}

With the insights described in Section~\ref{sec:background}, we now present our framework on inference-time scaling for diffusion models. We formulate the challenge as a search problem over the sampling noises; in particular, how do we know which sampling noises are good, and how do we search for such noises?

On a high-level, there are two design axes we propose to consider:
\begin{itemize}
    \item \textbf{Verifiers} are used to evaluate the goodness of candidates (Section~\ref{sec:verifiers}). These typically are some pre-trained models that are capable of providing feedback; concretely, verifiers are functions \begin{align}
        \mathcal{V}: \mathbb{R}^{H \times W \times C} \times \mathbb{R}^d \to \mathbb{R}
    \end{align}
    that takes in the generated samples and optionally the corresponding conditions, and outputs a scalar value as the score for each generated sample.
    \item \textbf{Algorithms} are used to find better candidates based on the verifiers scores (Section~\ref{sec:algorithm}). Formally defined, algorithms are functions \begin{align}
        f: \mathcal{V} \times D_\theta \times \{\mathbb{R}^{H \times W \times C} \times \mathbb{R}^d\}^N \to \mathbb{R}^{H\times W \times C}
    \end{align}
    that takes in a verifier $\mathcal{V}$, a pre-trained Diffusion Model $D_\theta$, and $N$ pairs of generated samples and corresponding conditions, and outputs the best initial noises according to the deterministic mapping between noises and samples. Throughout this search procedure, $f$ typically performs multiple forward passes through $D_\theta$ (see Section~\ref{sec:algorithm}). We refer to these additional forward passes as the \emph{search cost}, which we measure in terms of NFEs as well.
\end{itemize}

To give a concrete instantiation of our framework, we present a design walk-through of class-conditional ImageNet~\citep{deng2009imagenet} generation task. We take a SiT-XL~\citep{ma2024sit} model pre-trained on ImageNet with resolution of $256\times256$ and perform sampling with a second-order Heun sampler~\citep{karras2022elucidating}, \emph{i.e.}, no other source of randomness but the initial noise used in sampling. We measure inference compute budget with the total NFEs used with \textit{denoising steps} and \textit{search cost}. The denoising steps is fixed to the optimal setting of 250 NFEs~\citep{ma2024sit}, and we primarily investigate the scaling behavior with respect to the NFEs devoted to search. Unless specified otherwise, we use classifier-free guidance (cfg)~\citep{ho2022classifier} weight of $1.0$, focusing on the simple conditional generation task without guidance.

We start with the simplest search algorithm, where we randomly sample Gaussian noises, generate samples from them with ODE, and select those that correspond to the best verifier score~(Figure~\ref{fig:overview}). We denote such algorithm \textbf{Random Search}, which is essentially a Best-of-N strategy applied once on all noise candidates. Here, the primary axis for scaling NFEs in search is simply the number of noise candidates to select from. For verifiers, we start with the ``best'' one, an \textbf{Oracle Verifier}, which we assume to have full privileged information about the final evaluation of the selected samples. For ImageNet, since FID~\citep{heusel2017gans} and IS~\citep{salimans2016improved} are typically used as evaluation metrics, we directly take them as the oracle verifiers.

For IS, we select the samples with the highest classification probability output by a pre-trained InceptionV3 model~\citep{szegedy2016rethinking} of the conditioning class. For FID, we use the pre-calculated ImageNet Inception feature statistics (mean and covariance) as references, and we greedily choose the sample that minimizes the divergence against the ground-truth statistics. More details are included in Appendix~\ref{app:sec:exp-setting}.

\begin{figure}[t]
  \centering
  \begin{minipage}[b]{0.48\linewidth}
        \includegraphics[width=\linewidth]{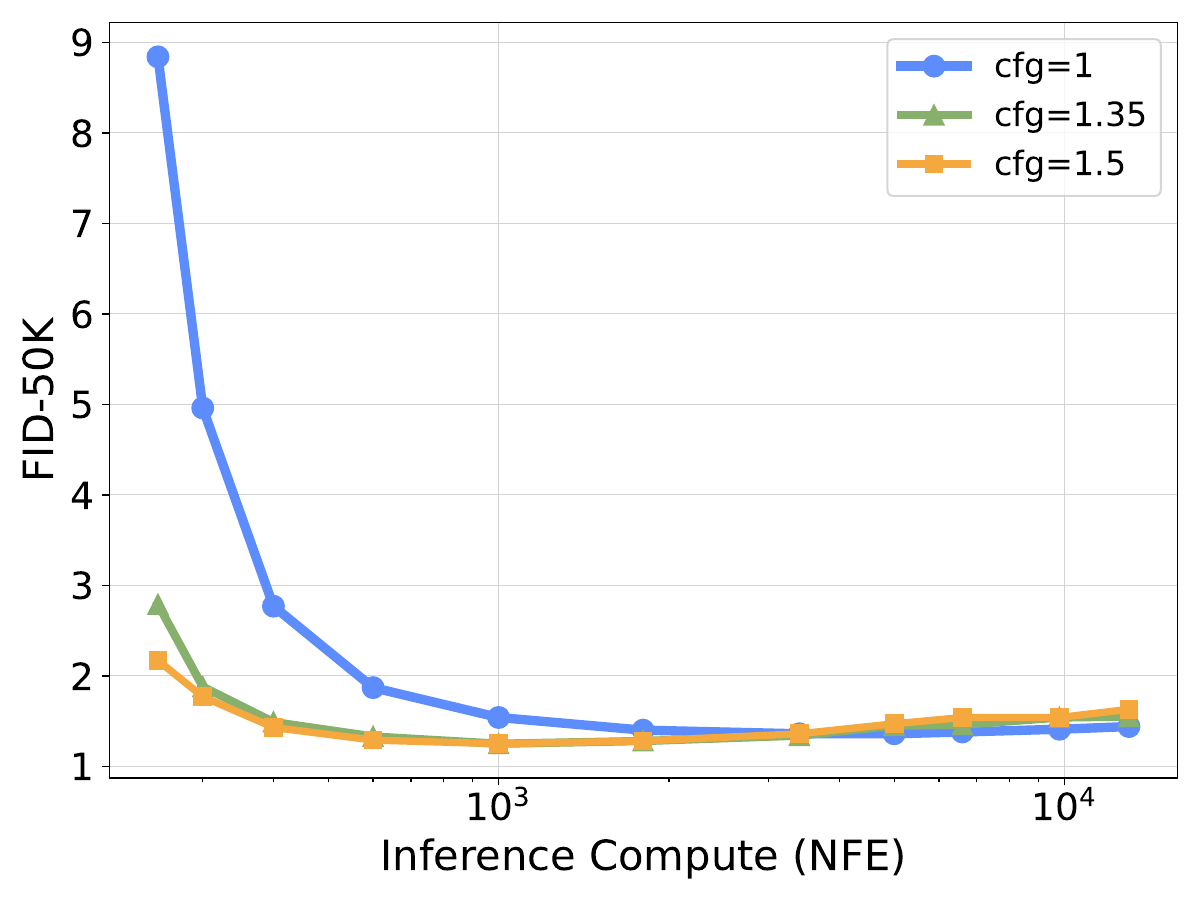}
    \end{minipage}
    \begin{minipage}[b]{0.48\linewidth}
        \includegraphics[width=\linewidth]{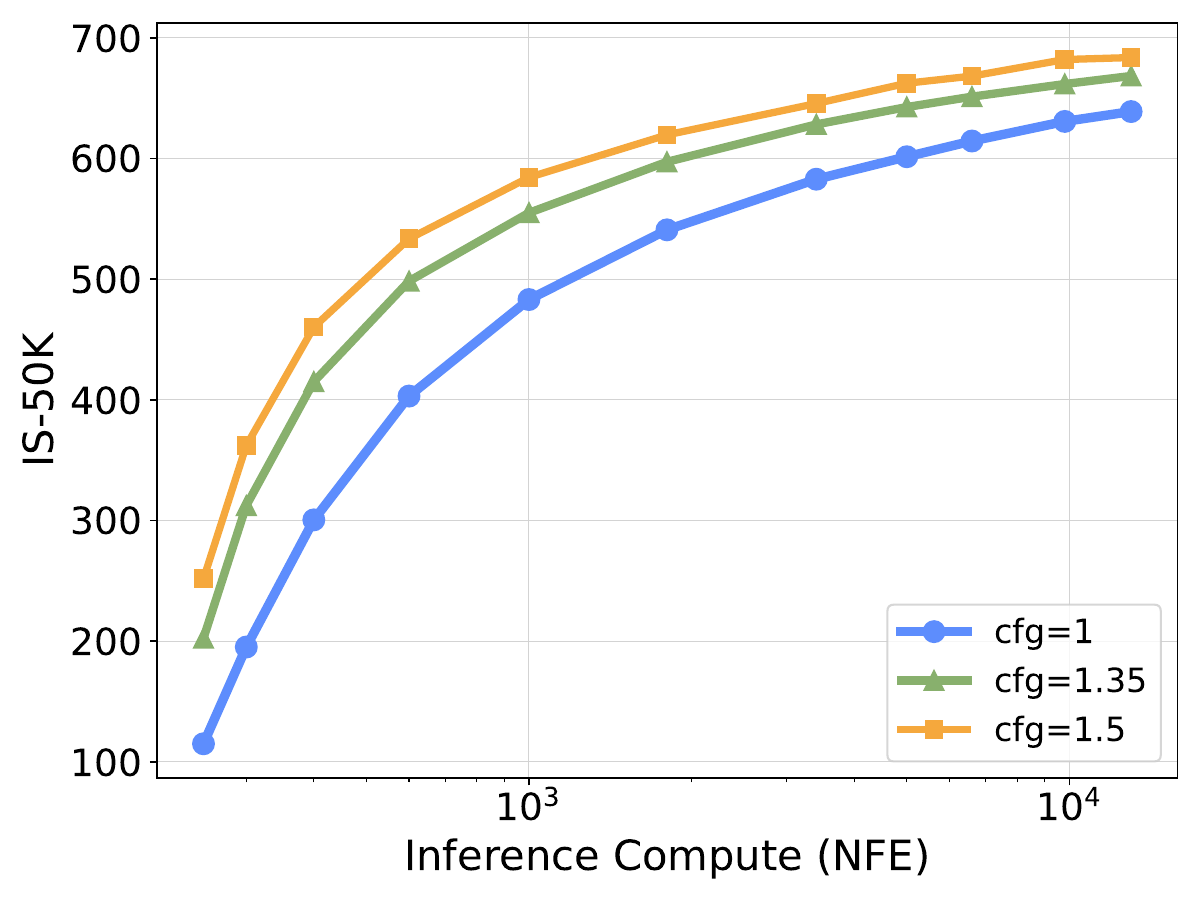}
    \end{minipage}
  
  \caption{\footnotesize{\textbf{\emph{Performances of Oracle Verifiers.}} Random Search with FID and IS on ImageNet. \textbf{Inference Compute} is given by the total NFEs devoted to denoising steps and search; the starting points of all curves in each and the following figures denote only devoting NFEs to denoising steps and 0 NFEs in search.}}
  \label{fig:BoN-FID-IS}
  \vspace{-0.5cm}
\end{figure}

As shown in Figure~\ref{fig:BoN-FID-IS}, the straightforward strategy in Random Search is highly effective across all guidance weights. As the NFEs invested in search increases, both FID and IS enjoy substantial improvements with their corresponding oracle verifiers. However, it is important to point out that in most cases it is impractical to directly employ the oracle verifier, since the specifics of the final evaluation procedures are generally not available. Therefore, this setting and the results are merely proof-of-concept, which serves as a confirmation that it is possible to invest compute into search and scale significantly at inference time, provided that the verifiers are chosen appropriately. 

\subsection{Search Verifiers}
\label{sec:verifiers}

In more realistic setups, verifiers could have access to the conditioning used for generation and some pre-trained models that are not explicitly aligned with the final evaluation procedures. In this scenario, the verifiers would evaluate the candidates based on both the quality of the samples and their alignment with the specified conditioning inputs. We denote such family the \textbf{Supervised Verifiers}.

While scaling NFEs with search demonstrates impressive performance with the oracle verifiers as displayed in Figure~\ref{fig:BoN-FID-IS}, the key question is whether its effectiveness can be generalized to supervised verifiers with more accessible pre-trained models designed for various vision tasks. To investigate this, we take two models with good learned representations, CLIP~\citep{radford2021learning} and DINO~\citep{oquab2023dinov2}. Since we only have class labels as the conditioning information on ImageNet, we utilize the classification perspective of the two models. For CLIP, we follow \citet{radford2021learning} and use the embedding weight generated via prompt engineering$^3$ as a zero-shot classifier. For DINO, we directly take the pre-trained linear classification head. During search, we run samples through the classifiers and select the ones with the highest logits corresponding to the class labels used in generation. We include more settings in Appendix~\ref{app:sec:exp-setting}.

\begin{wrapfigure}{l}{0.5\textwidth}
    \vspace{-0.8cm}
  \begin{center}
      \includegraphics[width=0.46\textwidth]{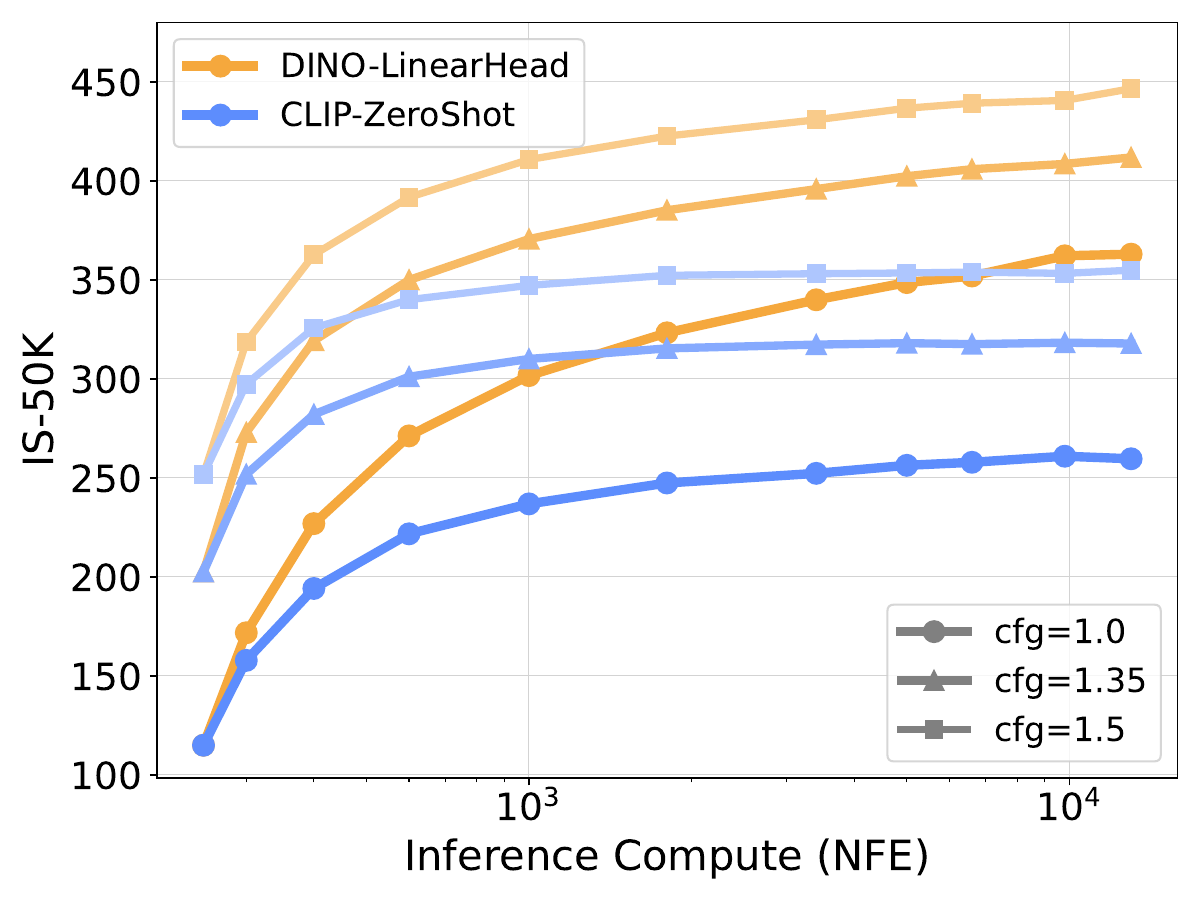}
  \end{center}
  \vspace{-0.5cm}
  \caption{\footnotesize{\textbf{\emph{Performances of Supervised Verifiers.}} Random Search with CLIP and DINO on ImageNet across different classifier-free guidance weights. \textbf{CLIP-ZeroShot} refers to using the logits output by the CLIP zero-shot classifier formulated with Prompt Engineering, and \textbf{DINO-LinearHead} refers to using the pre-trained linear classifier provided by~\citet{oquab2023dinov2}}.}
  \label{fig:search-classifier}
  \vspace{-0.5cm}
\end{wrapfigure}

As shown in Figure~\ref{fig:search-classifier}, this strategy also effectively improves the model performance on IS compared to the purely scaling NFEs with increased denoising steps (Figure~\ref{fig:teaser}). Nevertheless, we note that, as these classifiers operate point-wise, they are only partially aligned with the goal of FID score (see Appendix~\ref{app:sec:fid-loss-diversity}). Specifically, the logits they produce only focus on the quality of a single sample without taking population diversity into consideration, which leads to a significant reduction in sample variance and eventually manifests as mode collapse as the compute increases. The random search algorithm is also to blame due to its unconstrained search space, which accelerates the converging of search towards the bias of verifiers. Such phenomenon is similar to \emph{reward hacking} in reinforcement learning~\citep{pan2022effects, clark2024directly}, and thus we term it as \emph{Verifier Hacking}.

\fancypagestyle{footnote3}{\fancyfoot[L]{\footerfont $^3$\textit{See} \href{https://github.com/openai/CLIP/blob/main/notebooks/Prompt_Engineering_for_ImageNet.ipynb}{
\texttt{https://github.com/openai/CLIP/blob/main/notebooks/Prompt\_Engineering\_for\_ImageNet.ipynb}
}}}
\thispagestyle{footnote3}

\begin{wrapfigure}{r}{0.5\textwidth}
    \centering
    \begin{minipage}[b]{0.38\linewidth}
        \centering
        \includegraphics[width=\linewidth]{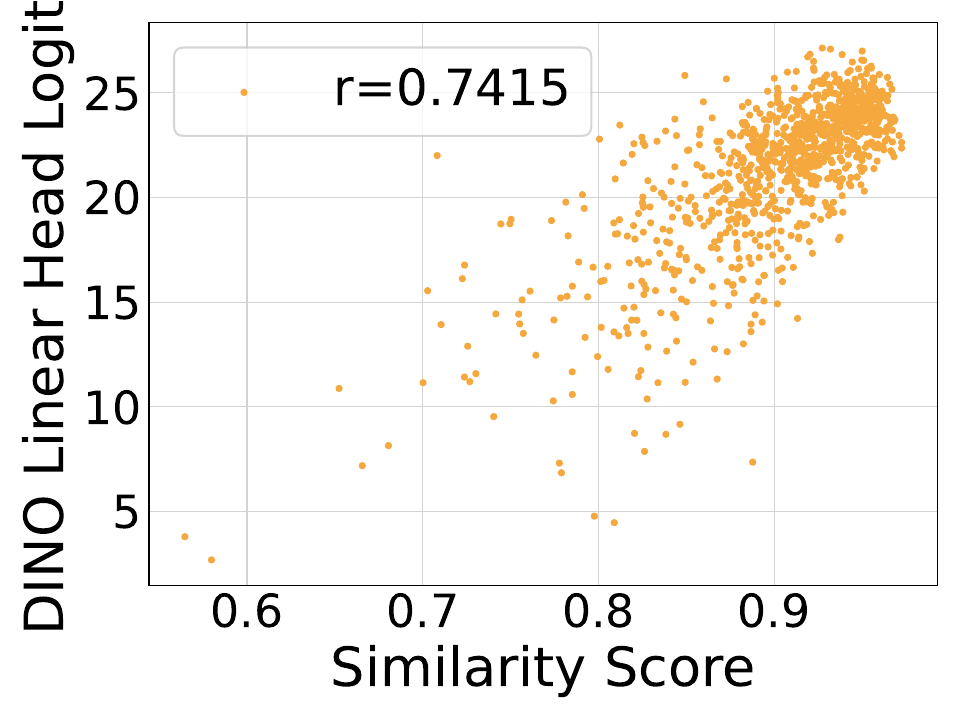}
        \includegraphics[width=\linewidth]{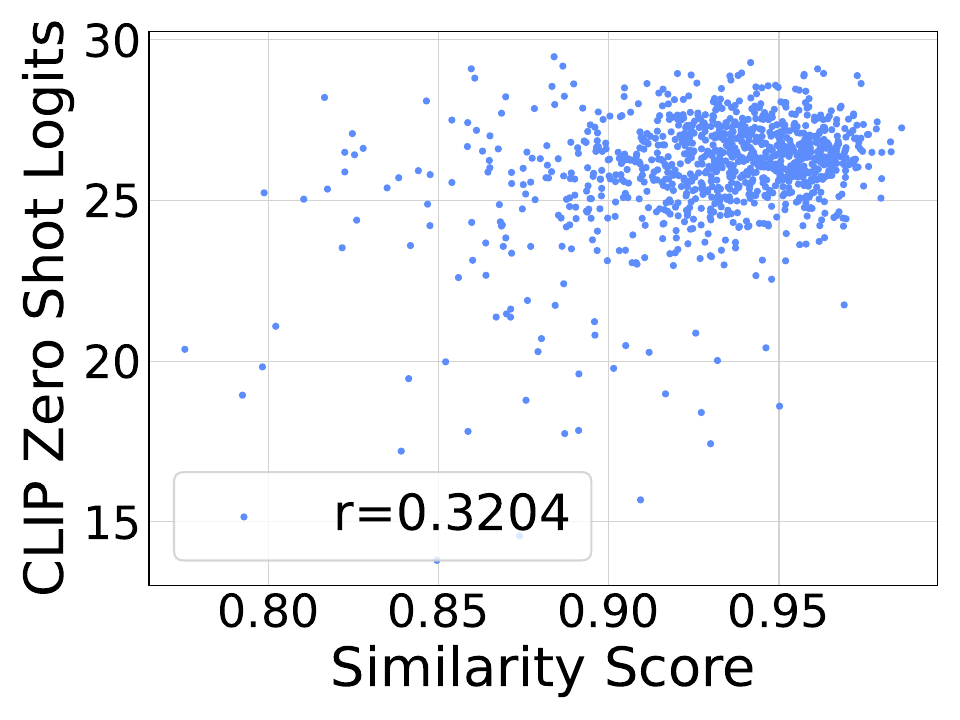}
    \end{minipage}
    \begin{minipage}[b]{0.6\linewidth}
        \includegraphics[width=\linewidth]{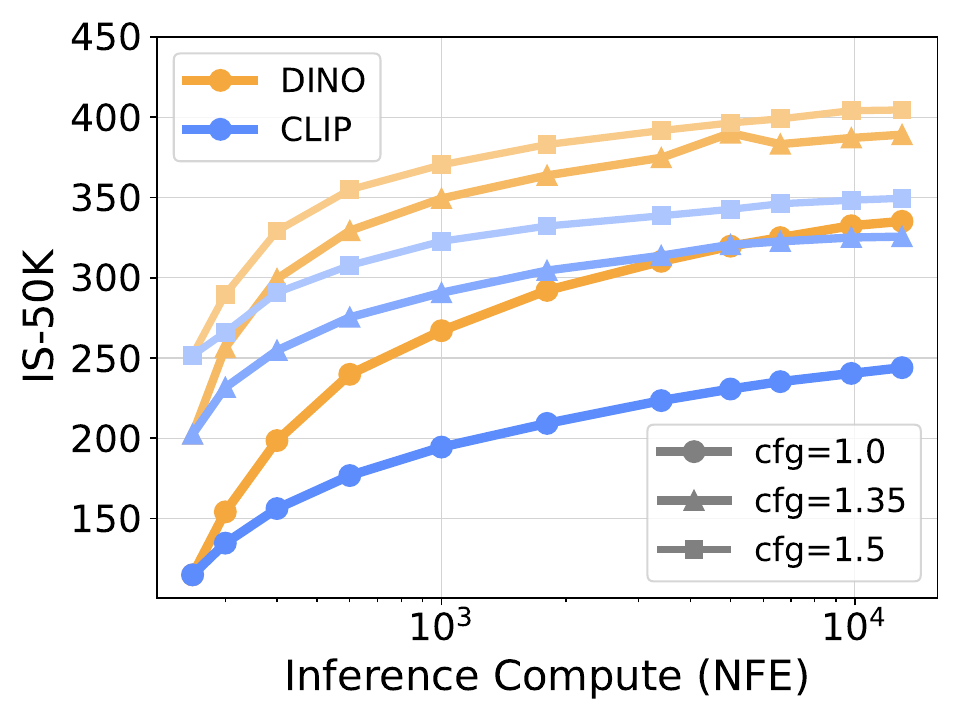}
    \end{minipage}
    
    \caption{\footnotesize{\textbf{\emph{Performances of Self-Supervised Verifiers.}} \textbf{Left}: correlation between CLIP and DINO feature similarity score and their classification logits; \textbf{Right}: Random Search with CLIP and DINO feature similarity score as verifiers across different classifier-free guidance weight.}}
    \label{fig:search-ssl}
    \vspace{-0.5cm}
\end{wrapfigure}

Although conditioning information is essential in real-world generation tasks, we discover that it is not necessary for the verifiers to guide the search effectively. As shown in Figure~\ref{fig:search-ssl}, we find that there exists a strong correlation between the logits output by the DINO / CLIP classifiers, and the feature space (extracted by DINO / CLIP, respectively) cosine similarity of the model's x-prediction at a low noise level ($\sigma = 0.4$) and the final generated clean sample ($\sigma = 0$). Therefore, we proceed to use this similarity score as a surrogate for classification logits, and denote such family \textbf{Self-Supervised Verifiers}, given that they do not require extra condition information. We again observe effective scaling behavior in Figure~\ref{fig:search-ssl}. This result is encouraging for use cases where conditioning information is not available or hard to obtain, like the task of medical imaging generation~\citep{khader2022medical}. As these limitations are uncommon in real-world settings, we leave further investigation of Self-Supervised Verifiers to future work.

\vspace{-0.2cm}
\subsection{Search Algorithms} 
\vspace{-0.2cm}
\label{sec:algorithm}

Our previous explorations have predominantly considered a simple random search setup, which is a one-time best-of-N selection strategy on a randomly chosen fixed set of candidates. Our findings in Section~\ref{sec:verifiers} indicate that this approach can lead to \textit{verifier hacking}: since the random search operates on the entire Gaussian space, it can quickly overfit to the ``bias'' of verifiers and lead to failure of our intended goal~\citep{gao2023scaling}. This realization motivates us to investigate more nuanced search algorithms that leverage verifiers' feedback to iteratively refine candidates, only slightly each time, thus mitigating the overfitting risks. Specifically, we consider a \textbf{Zero-Order Search} approach:

\begin{enumerate}
    \item we start with a random Gaussian noise $\myvec{n}$ as pivot.
    \item find $N$ candidates in the pivot's neighborhood. Formally, the neighborhood is defined as $S_{\myvec{n}}^\lambda = \{\myvec{y}: d(\myvec{y},\myvec{n}) = \lambda\}$, where $d(\cdot, \cdot)$ is some distance metric.
    \item run candidates through an ODE solver to obtain samples and their corresponding verifier scores.
    \item find the best candidates, update it to be the pivot, and repeat steps 1-3.
\end{enumerate}

Much like Zero-Order optimization~\citep{flaxman2004online}, Zero-Order Search does not involve expensive gradient calculation; instead, it approximates the gradient direction via multiple forward function evaluations inside the neighborhood. As with standard first-order methods~\citep{ruder2016overview}, we deem the number of iterations (i.e., how many times the algorithm runs through steps 1-3) to be the primary axis for scaling NFEs in search. When $N$ gets larger, the algorithm will locate a more precise local ``optimum'', and when $\lambda$ increases, the algorithm will have a larger stride and thus traversing the noise space more quickly. In practice, we fix the value of $\lambda$ and investigate the scaling behavior w.r.t $N$. We abbreviate the algorithm as \textbf{ZO-$N$}.

We note that since many verifiers are differentiable, first-order search with true gradient is technically possible and has seen applications in practice~\citep{novack2024ditto,ben2024d}. However, it requires back-propagating through the entirety of the sampling process, which is typically prohibitively costly in terms of both time and space complexity, especially when scaling large models. In practice, we find that first-order search does not demonstrate a significant advantage over zero-order search on ImageNet despite its higher cost and worse scalability. We include the comparisons in Appendix~\ref{app:sec:first-order}.

The iterative nature of diffusion sampling processes yields other possibilities for designing local search algorithms, and it is feasible to search along the sampling trajectories over the noises injected. We propose \textbf{Search over Paths} to explore one of such possibilities. Specifically, 
\begin{enumerate}
    \item sample $N$ initial i.i.d. noises and run the ODE solver until some noise level $\sigma$. The noisy samples $\myvec{x}_\sigma$ serve as the search starting point.
    \item sample $M$ i.i.d noises for each noisy samples, and simulate the forward noising process from $\sigma$ to $\sigma + \Delta f$ to produce $\{\myvec{x}_{\sigma+\Delta f}\}$ with size $M$.
    \item run ODE solver on each $\myvec{x}_{\sigma+\Delta f}$ to noise level $\sigma+\Delta f-\Delta b$, and obtain $\myvec{x}_{\sigma+\Delta f-\Delta b}$. Run verifiers on these samples and keep the top $N$ candidates. Repeat steps 2-3 until the ODE solver reaches $\sigma=0$
    \item run the remaining $N$ samples through random search and keep the best one.
\end{enumerate}

To ensure the iteration terminates, we strictly require $\Delta b > \Delta f$. Also, since the verifiers are typically not adapted to noisy input, we perform one additional denoising step in step 3 and use the clean x-prediction to interact with the verifiers. Here, the primary scaling axis is the number of noises $M$ added in step $2$, and in practice, we investigate the scaling behavior with different numbers of initial noises $N$. We thus term the algorithm \textbf{Paths}-$N$. Both algorithms are illustrated in Figure~\ref{fig:overview}, from which we can see that compared to Random Search, both Zero-Order Search and Search over Paths retain very strong locality: the former operates in the neighborhood of the initial noise, and the latter searches in the intermediate steps of the sampling process.

\begin{figure}[t!]
    \centering
    \begin{minipage}[b]{0.48\linewidth}
        \centering
        \includegraphics[width=0.48\linewidth]{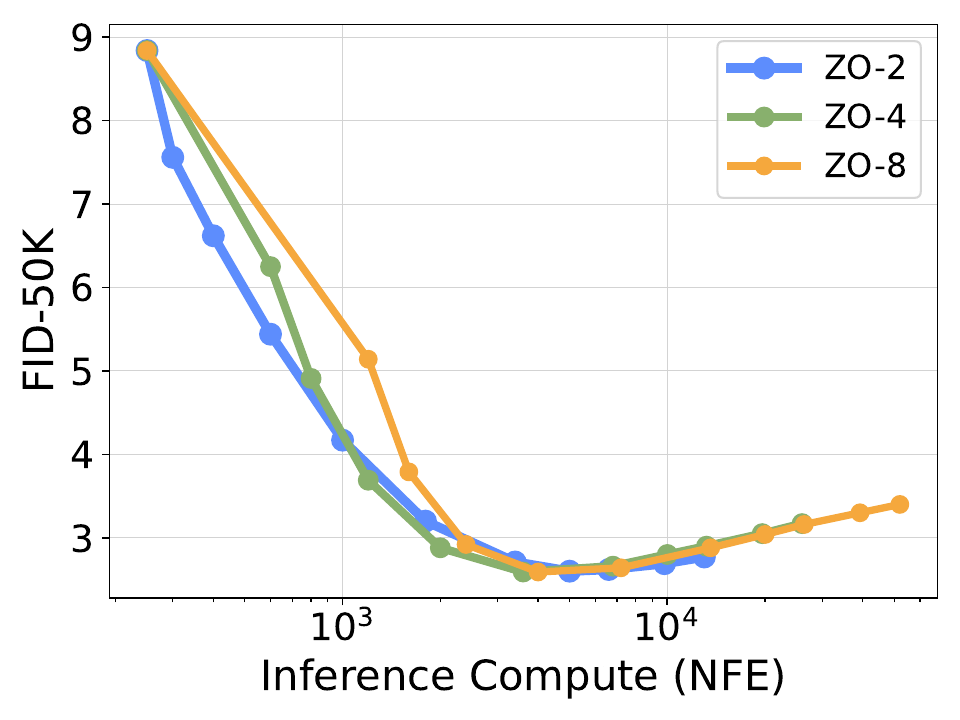}
        \includegraphics[width=0.48\linewidth]{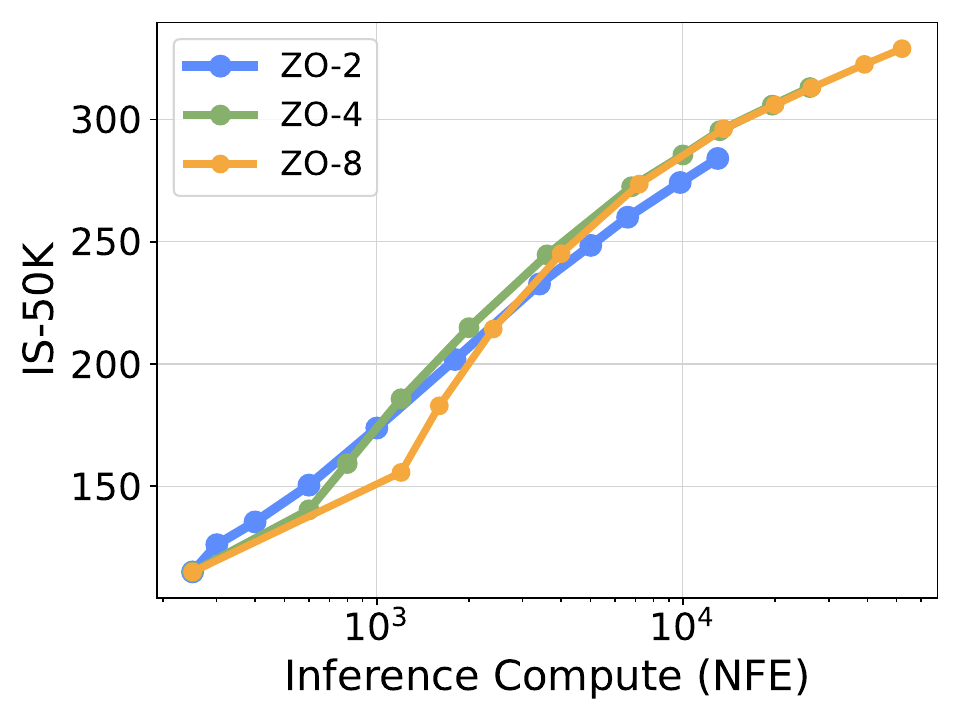}
    \end{minipage}
    \begin{minipage}[b]{0.48\linewidth}
        \centering
        \includegraphics[width=0.48\linewidth]{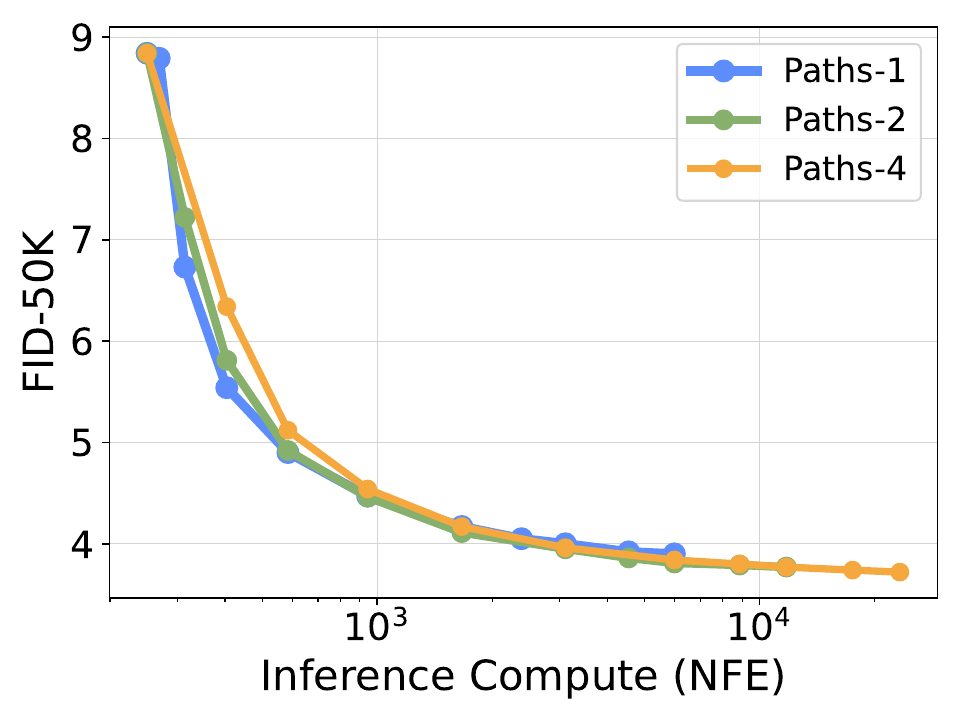}
        \includegraphics[width=0.48\linewidth]{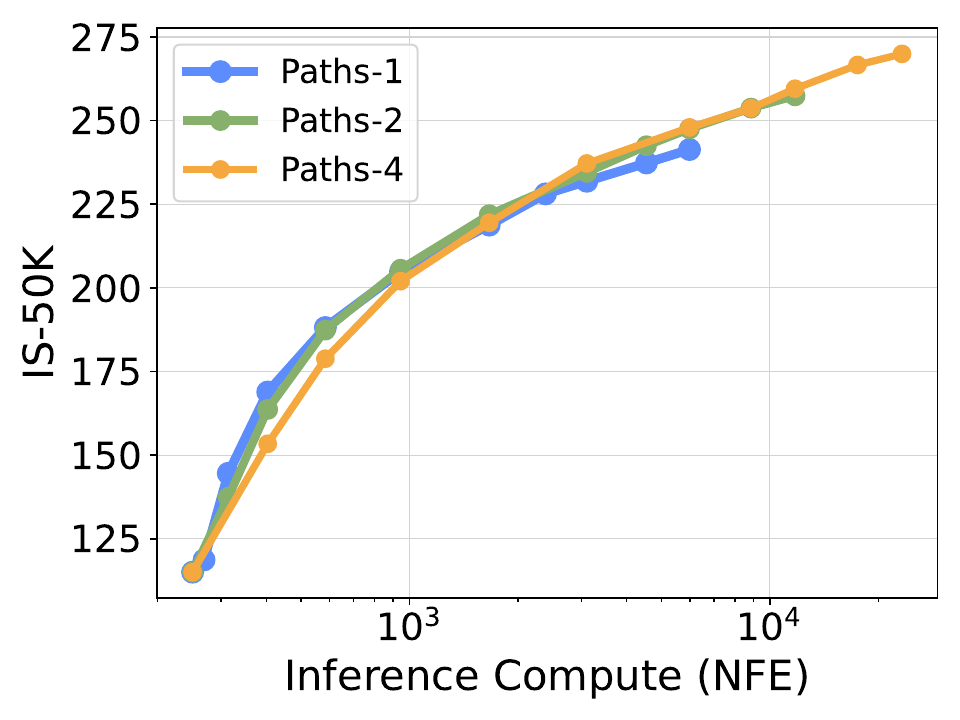}
    \end{minipage}
    
    \caption{\footnotesize{\textbf{\emph{Performances of Search Algorithms.}} We fix the verifier to be DINO-LinearHead and investigate the FID and IS of Zero-Order Search and Search over Paths on ImageNet. For each algorithm, we further demonstrate the relationship between $N$ and their performances.}}
    \label{fig:search-algos}
    \vspace{-0.5cm}
\end{figure}

We show the performance of these algorithms in Figure~\ref{fig:search-algos}. Due to the locality nature of the two algorithms, both of them manage to alleviate the diversity issue of FID to some extent while maintaining a scaling Inception Score. For Zero-Order Search, we note that the effectiveness of increasing $N$ is marginal, and $N=4$ seems to already be a good estimation of the local optimum. For Search over Paths, we see that different values of $N$ lead to different scaling behavior, with small $N$ being compute efficient in small generation budget, and large $N$ having an advantage when scaling up compute more.

\vspace{-0.2cm}
\section{Inference-Time Scaling in Text-to-Image}
\label{sec:t2i}
\vspace{-0.2cm}

With the instantiation of our search framework in Section~\ref{sec:imagenet}, we proceed to examine its inference-time scaling capability in larger-scale text-conditioned generation tasks, and study the alignment between verifiers and specific image generation tasks.

\textbf{Datasets.} For a more holistic evaluation of our framework, we use two datasets: (1) DrawBench, introduced in~\citet{saharia2022photorealistic}, consists of $200$ prompts spanning 11 different categories. It aims to evaluate text-to-image models' ability to handle complex prompts and generate realistic and high-quality images. During evaluations, we generate one image per prompt. (2) T2I-CompBench~\citep{huang2023t2i} is a benchmark designed for evaluating attribute binding, object relationships, and complex compositions. We generate two images per prompt and use the $1800$ prompts from the validation set for evaluation.

\textbf{Models.} We use the newly released FLUX.1-dev model~\citep{flux1dev} as our backbone, which is currently at the frontier of text-to-image generation and representative of the capabilities of many contemporary text-conditioned diffusion models. For detailed sampling settings, see Appendix~\ref{app:sec:exp-setting}.

\begin{figure}
    \centering
    \vspace{-0.5cm}
        \begin{overpic}[width=0.95\linewidth]{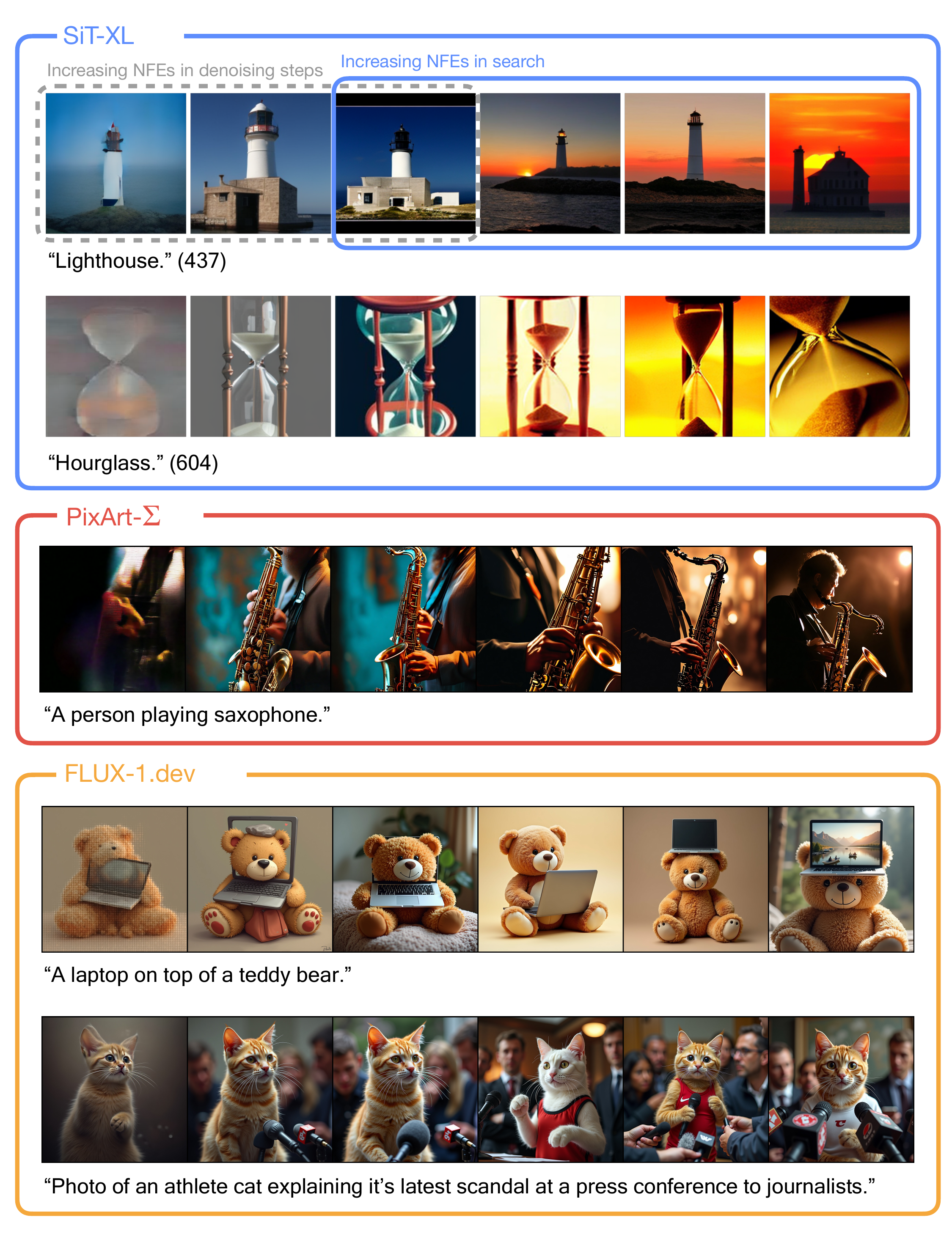}
            \put(16,36.7){\citep{flux1dev}}
            \put(13.4,57.7){\citep{chen2024pixart}}
            \put(11.1,96.7){\citep{ma2024sit}}
        \end{overpic}
    \vspace{-0.4cm}
    \caption{\footnotesize{\textbf{\emph{Visualizations of Scaling Behaviors.}} Each row is constructed as follows: \textbf{left three}: sampled with increasing NFEs in denoising steps; \textbf{right four}: sampled with increasing NFEs in search. First two rows are sampled from SiT-XL~\citep{ma2024sit} with DINO-LinearHead, third row is sampled from PixArt-$\Sigma$~\citep{chen2024pixart} with Verifier Ensemble, and last two rows are sampled from FLUX-1.dev~\citep{flux1dev} with Verifier Ensemble.}}
    \label{fig:visual-teaser}
\end{figure}

\textbf{Verifiers.} Due to the inherently sophisticated nature of text-conditioned image generation, a more comprehensive and fine-grained evaluation is required~\citep{lee2024holistic}. We therefore expand the choice of supervised verifiers to assess a broader range of aspects in the generated images: \textit{Aesthetic Score Predictor}$^4$~\citep{schuhmann2022laion}, \textit{CLIPScore}~\citep{hessel2021clipscore}, and \textit{ImageReward}~\citep{xu2024imagereward}. Relying on a large amount of human-annotated data, these verifiers capture human preferences from different perspectives: \textit{Aesthetic Score Predictor} is trained to predict the human rating of synthesized images' visual quality; \textit{CLIPScore} aligns visual and text features via 400M human labeled (image, text) pair data; and lastly, \textit{ImageReward} learns to capture more general preferences via carefully curated annotation pipeline, including rating and ranking samples on text-image alignment, aesthetic quality, and harmlessness. Therefore, \textit{ImageReward} has the larger capacity and can capture the evaluative aspects of \textit{Aesthetic Score} and \textit{CLIPScore} to some extent. We include more discussion and results in Section~\ref{sec:t2i-analysis}.

\fancypagestyle{footnote4}{
\fancyfoot[L]{\footerfont 
$^4$\textit{Though not taking in condition information, the aesthetic predictor is considered to be supervised by the annotated aesthetic scores on LAION}.\\
$^5$\textit{See model details on} \url{https://ai.google.dev/gemini-api/docs/models/gemini\#gemini-1.5-flash}}}
\thispagestyle{footnote4}

Additionally, we combine these three verifiers to create a fourth verifier, referred to as the \textit{Verifier Ensemble}, to further expand the capacity of verifiers across the evaluative aspects. Since the metrics produced by these verifiers operate on substantially different scales, instead of the absolute scores, we record the relative rankings of metrics across samples, configure the Verifier Ensemble to assess the unweighted average ranking of the three metrics for each sample, and select the sample with the highest ranking.

We find that self-supervised verifiers are less effective in text-to-image settings. We attribute this to two main factors: (1) self-supervised verifiers focus on the visual quality of images but overlook essential textual information, and (2) the large-scale pre-training and extensive fine-tuning might make text-to-image models attain very different sampling dynamics compared to small class-conditioned models trained on ImageNet. We include the performance and more detailed analysis in Appendix~\ref{app:sec:t2i-ssl}.

\textbf{Metrics.} On DrawBench, we use all verifiers not employed in the search process as primary metrics to provide a more comprehensive evaluation. Considering the usage of Verifier Ensemble, we additionally introduce an LLM as a neutral evaluator for assessing sample qualities.

The extensive pretraining and substantial model capacity of LLMs and Multimodal Large Language Models (MLLM) endow them with exceptional image-text understanding and generalization capabilities, making them highly effective evaluators for assessing the quality of synthesized images across diverse aspects~\citep{tan2024evalalign}. In fact, many prior works either adopt the VQA approach with LLMs as evaluation models~\citep{hu2023tifa,wiles2024revisiting,lu2024llmscore}, or leverage MLLMs as annotators to obtain feedback on various aspects of the generated images~\citep{wu2024multimodal,chen2024mj}. Inspired by these approaches, we prompt the Gemini-1.5 flash model (via \texttt{Gemini-1.5-Flash-002} API$^5$) to assess synthesized images from five different perspectives: Accuracy to Prompt, Originality, Visual Quality, Internal Consistency, and Emotional Resonance. Each perspective is rated on a scale from $0$ to $100$, and the averaged overall score is used as the final metric. We denote such evaluator as \textit{LLM Grader}, and include the prompting and evaluation setup in Appendix~\ref{app:sec:exp-setting}.

Lastly, on T2I-CompBench, we use the evaluation pipeline provided by~\citet{huang2023t2i} to assess the performance of our framework in compositional generation tasks. The pipeline utilizes the BLIP-VQA model~\citep{li2022blip} for attribute binding evaluation, UniDet model~\citep{zhou2022simple} for spatial relationship evaluation, and finally weighted average of BLIP, UniDet, and CLIP Score for evaluating complex compositions.

\vspace{-0.25cm}
\subsection{Analysis Results: Verifier-Task Alignment}
\label{sec:t2i-analysis}

We now present our results comparing combinations of verifiers and algorithms on different datasets.

\begin{figure}[t]
    \centering
    \includegraphics[width=0.9\linewidth]{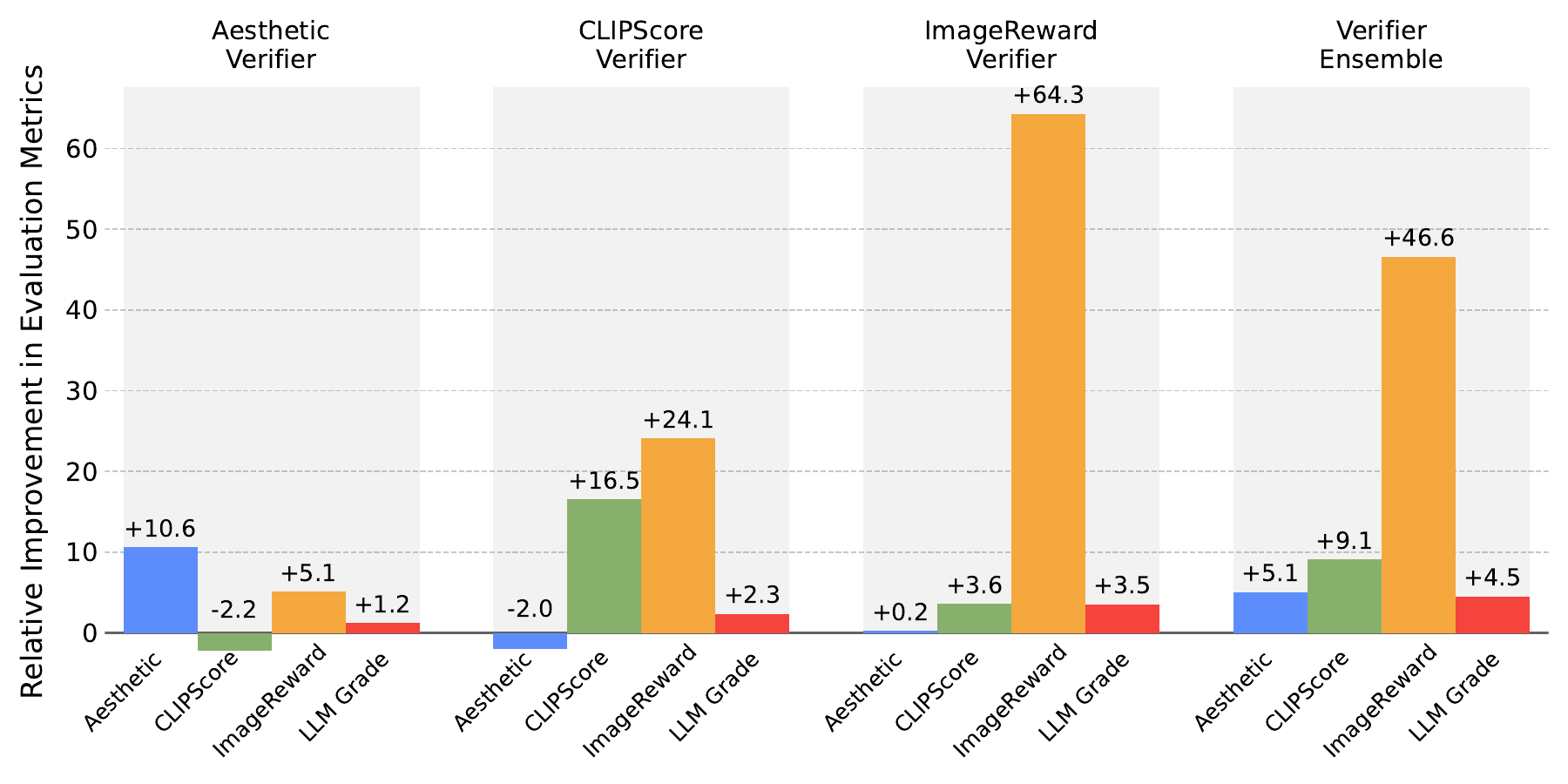}
    
    \caption{\footnotesize{\textbf{\emph{Performances of Search with FLUX.1-dev at inference-time.}} We fix the search budget to be $3840$ NFEs with random search, and demonstrate the relative performance gain (\%) with generation without any search budget.}}
    \label{fig:flux-metrics}
    \vspace{-0.5cm}
\end{figure}

\begin{wrapfigure}{l}{0.5\textwidth}
    \vspace{-0.5cm}
    \centering
    \includegraphics[width=0.45\textwidth]{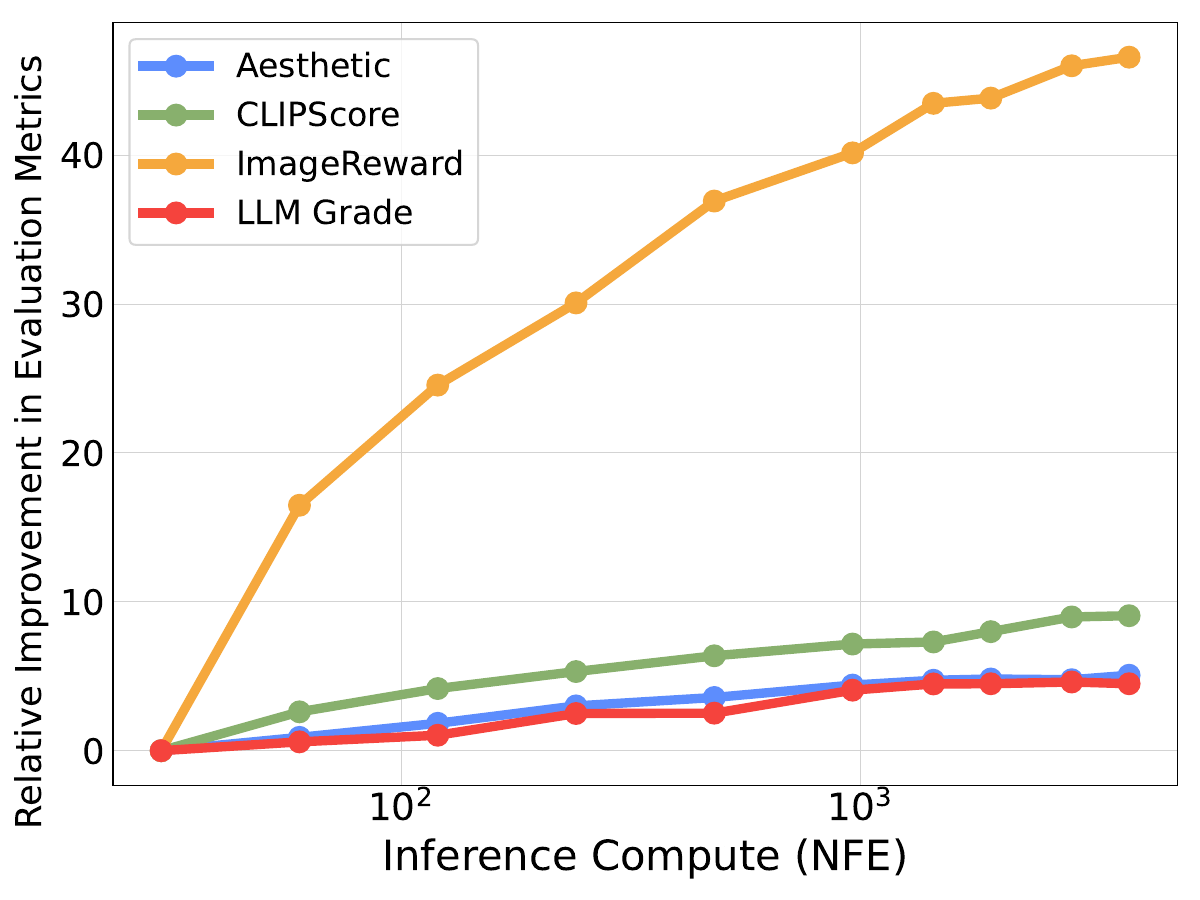}
    \caption{\footnotesize{\textbf{\emph{Scalability of search with FLUX.1-dev on DrawBench.}} We use random search with Verifier Ensemble to obtain the results, and demonstrate the relative performance gain (\%) with generation without any search budget. Similar scaling behavior to ImageNet setting is observed across different metrics.}}
    \label{fig:random-search-ensemble}
    \vspace{-0.5cm}
\end{wrapfigure}

\textbf{DrawBench.} DrawBench is a highly general-purpose dataset containing text prompts from diverse categories. We argue that evaluating generation tasks on such dataset requires assessing a wide range of aspects rather than focusing narrowly on specific criteria (e.g., aesthetic, text alignment). Given its comprehensive pretraining knowledge and the diverse evaluative aspects we established, the LLM Grader serves as an effective surrogate for human preferences on this benchmark. By leveraging it, we can assess how well a verifier aligns with the broad requirements of generation tasks on DrawBench.

As illustrated in Figure~\ref{fig:flux-metrics}, and as indicated by the LLM Grader, searching with all verifiers generally improves sample quality, while specific improvement behaviors vary across different setups. This demonstrates our claim that search setups can be specifically chosen to conform to different application scenarios. For instance, ImageReward and Verifier Ensemble, which possess more nuanced evaluative aspects and closely align with human preferences, consistently improve scores across all evaluation metrics, making them suitable for the generalized generation tasks on DrawBench. In contrast, Aesthetic and CLIP Verifiers are less desirable for tasks requiring satisfying performances across multiple evaluative aspects, due to the effect of \textit{verifier hacking}.

From the left two columns in Figure~\ref{fig:flux-metrics}, we see that searching with Aesthetic and CLIP Verifier overfit to their inherent biases, negatively impacting each other. We conjecture that both verifiers suffer from major misalignment in evaluation: the Aesthetic Score focuses solely on visual quality, often favoring highly stylized images deviating from their text prompt, whereas CLIP prioritizes visual-text alignment at the expense of visual quality~\citep{clark2024directly, wallace2024diffusion, yang2024using}. As a result, exploiting the biases of one verifier (e.g. Aesthetic Score) during search degrades the evaluation metrics assessed by the other verifier (e.g. CLIP). This aligns with observations by~\citet{clark2024directly}, who noted that extensive fine-tuning with Aesthetic or CLIP rewards can cause the model distribution to collapse to a single high-reward mode.

However, we point out that our search method does not modify the model's learned score function and will preserve its pretrained behavior on individual samples. Consequently, unlike the complete collapsing in single-sample quality observed by~\citet{clark2024directly}, the samples selected by our search method remain within the learned data distribution, only with their mode shifted towards one of the verifiers (say Aesthetic) and slightly away (only $\sim-2\%$ in performance) from the other (say CLIP). This is further supported by the evaluation results from the LLM Grader, that searching for Aesthetic or CLIP Score can still improve the overall preference scores, despite trade-offs between the two. Importantly, since searching with Aesthetic and CLIP Score does not lead to a total collapse in sample quality and leverages their unique strengths in aesthetic quality and text faithfulness, they can be well-suited for tasks that require a focus on specific attributes such as visual appeal or textual accuracy, rather than maintaining general-purpose performance.

Lastly, from Figure~\ref{fig:random-search-ensemble}, we observe similar scaling behavior of evaluation metrics with respect to increasing search budget, similar to the ImageNet settings.

\begin{wraptable}{r}{0.5\textwidth}
\centering
\scalebox{0.62}{
\begin{tabular}{lcccccc}
\toprule
\multicolumn{1}{c}{\textbf{Verifier}} & Color & Shape & Texture & Spatial & Numeracy & Complex \\
\midrule
- & 0.7692 & 0.5187 & 0.6287 & 0.2429 & 0.6167 & 0.3600 \\
Aesthetic & 0.7618 & 0.5119 & 0.5826 & 0.2593 & 0.6159 & 0.3472 \\
CLIP & 0.8009 & 0.5722 & 0.7005 & 0.2988 & 0.6457 & 0.3704 \\
ImageReward & \textbf{0.8303} & \textbf{0.6274} & \textbf{0.7364} & \textbf{0.3151} & \textbf{0.6789} & \textbf{0.3810} \\
\midrule
Ensemble & 0.8204 & 0.5959 & 0.7197 & 0.3043 & 0.6623 & 0.3754 \\
\bottomrule
\end{tabular}
}
\caption{\footnotesize{\textbf{\emph{Performance of search with FLUX.1-dev on T2I-CompBench.}} We use random search with Verifier Ensemble to obtain the samples; for evaluation, we use the pipeline provided in T2I-CompBench. The first row denotes the performance without search where we fix the denoising budget to be $30$ NFEs, and for the rest, we fix the search budget to be $1920$ NFEs.}}
\label{tab:comp-bench}
\vspace{-0.5cm}
\end{wraptable}

\textbf{T2I-CompBench.} Unlike DrawBench, the evaluation pipeline on T2I-CompBench primarily emphasize correctness in relation to the text prompt~\citep{huang2023t2i}, such as accurately generating colors, object relationships, and overall compositions, without prioritizing pure visual quality. These different goals effectively call for a different search setup, and the results from Table~\ref{tab:comp-bench} support this claim. We have observed that searching with Aesthetic Scores leads to minimal improvements and even degradation in metrics. Although all three remaining verifiers account for text faithfulness to some extent, they demonstrate varying degrees of improvement. Notably, ImageReward outperforms Verifier Ensemble across all evaluation categories, while CLIP provides only marginal gains. This can be attributed to the fact that CLIP lacks the nuanced evaluative aspects aligned with human preferences, and Verifier Ensemble includes Aesthetic Score, which negatively impacts evaluation performance on this task

These contrasting behaviors of verifiers on DrawBench and T2I-CompBench highlight that the effectiveness of a verifier depends on how well its criteria align with the specific requirements of the task, with certain verifiers being better suited for particular tasks than others. This inspires the design of more task-specific verifiers, which we leave as future works.

\textbf{Algorithms.} In Table~\ref{tab:algo-draw-bench} we demonstrate the performance of the three presented search algorithms on DrawBench. For Zero-Order Search, we set a fixed number for neighbors, $N = 2$. For Search over Paths, we set the number of initial noises, $N = 2$, as well. More detailed settings are included in Appendix~\ref{app:sec:exp-setting}.

We see that all three methods can effectively improve the sampling quality, with random search outperforming the other two methods in some aspects. Again we credit this behavior to the locality nature of Zero-Order Search and Search over Paths (Figure~\ref{fig:overview}). Since all verifiers and metrics we present are evaluated on a per-sample basis, random search will drastically accelerate the convergence to the bias of verifiers, whereas the other two algorithms need to perform refinement on the suboptimal candidates.

\begin{table}[t!]
\centering
\scalebox{0.85}
{

\begin{tabular}{rcccc}
\toprule

\multicolumn{1}{c}{\textbf{Verifier}} & Aesthetic & CLIPScore & ImageReward & LLM Grader \\
\midrule
- & 5.79 & 0.71 & 0.97 & 84.29 \\
\midrule
Aesthetic + Random & \textbf{6.38} & 0.69 & 0.99 & 86.04 \\
          + ZO-2 & 6.33 & 0.69 & 0.96 & 85.90\\
          + Paths-2 & 6.31 & 0.70	& 0.95 & 85.86\\
\midrule
CLIPScore + Random & 5.68	& \textbf{0.82} & 1.22 & 86.15 \\
          + ZO-2 & 5.72 &  0.81 & 1.16 & 85.48\\
          + Paths-2 & 5.71 & 0.81 & 1.14 & 85.45\\
\midrule
ImageReward + Random & 5.81 & 0.74 & \textbf{1.58} &  87.09 \\
            + ZO-2 & 5.79	& 0.73 & 1.50 & 86.22\\
            + Paths-2 & 5.76 & 0.74 & 1.49	& 86.33\\
\midrule
Ensemble + Random & 6.06 & 0.77 & 1.41 & \textbf{88.18}\\
         + ZO-2 & 5.99 & 0.77 & 1.38 & 87.25\\
         + Paths-2 & 6.02 & 0.76	& 1.34 & 86.84\\
\bottomrule
\end{tabular}
}
\caption{\footnotesize{\textbf{\emph{Performance of search algorithms with different verifiers on DrawBench.}} The results are obtained from FLUX.1-dev evaluated on DrawBench. The first row denotes the performance without search where we fix denoising budget to be $30$ NFEs, and for the rest we fix search budget to be $2880$ NFEs.}}
\label{tab:algo-draw-bench}
\vspace{-0.5cm}
\end{table}

\vspace{-0.5cm}
\subsection{Search is Compatible with Finetuning}
\label{sec:search-finetune}
\vspace{-0.2cm}

Both search and finetuning methods \citep{clark2024directly, fan2024reinforcement} aim to align the final samples with explicit reward models or human preferences. While the former shifts the sample modes toward the bias of specific verifiers, the latter directly modifies the model's distribution to align with the rewards. This raises a natural question: can we still shift the sample modes according to verifiers after the model distribution has been modified?

\begin{wraptable}{l}{0.5\textwidth}
\centering
\scalebox{0.8}{
\begin{tabular}{rcccc}
\toprule
\multicolumn{1}{c}{\textbf{Model}} & Aesthetic & CLIP & PickScore \\
\midrule
SDXL & 5.56 & 0.73 & 22.39  \\
\midrule
+ DPO & 5.59 & 0.74 & 22.54  \\
+ DPO \& Search & \textbf{5.66} & \textbf{0.76} & \textbf{23.54}  \\
\bottomrule
\end{tabular}
}
\caption{\footnotesize{\textbf{\emph{Performance of Search with DPO-finetuned SDXL.}} We use random search with Verifier Ensemble on DrawBench to obtain the result. We set the denoising budget to $40$ NFEs, and search budget to $1280$ NFEs.}}
\vspace{-0.5cm}
\label{tab:dpo-sdxl}
\end{wraptable}
Among all finetuning methods explored, Diffusion-DPO~\citep{wallace2024diffusion}, as a more efficient and simpler alternative to RLHF~\citep{ouyang2022training} methods, has been widely adopted in aligning large-scale text-to-image models. To answer the question, we take the DPO fine-tuned Stable Diffusion XL model in~\citep{wallace2024diffusion} and conduct search on the DrawBench dataset. Since the model is finetuned on the dataset Pick-a-Pic~\citep{kirstain2023pick}, we replace ImageReward with the PickScore evaluator. The results are included in Table~\ref{tab:dpo-sdxl}.

We see that search method can generalize to different models and can improve the performance of an already aligned model. We note this will be a useful tool to mitigate the cases where finetuned models disagree with reward models~\citep{hu2024new}, and to improve their generalizability to other metrics~\citep{clark2024directly}.

\section{Axes of Inference Compute Investment}
\label{sec:dimension}

Due to the iterative sampling nature of diffusion models, there are multiple dimensions in which we can scale NFEs with search. We present them below and investigate their impact on performances.

\textbf{Number of search iterations.} Intuitively, increasing the number of search iterations allows the selected noises to approach the optimal set with respect to verifiers and can thus substantially improve performance. We have observed such behavior in all of our previous experiments.

\begin{wrapfigure}{r}{0.5\textwidth}
    \vspace{-0.5cm}
    \centering
    \includegraphics[width=0.38\textwidth]{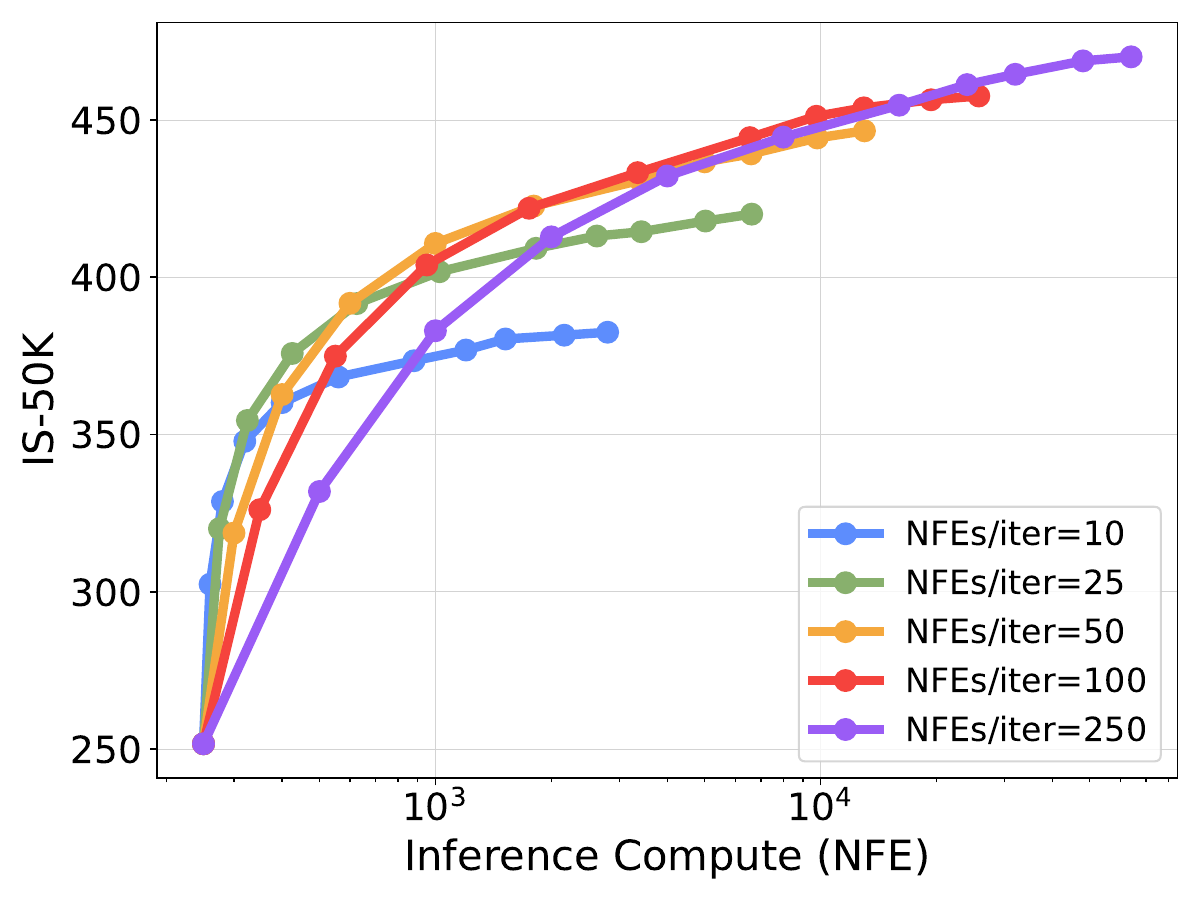}
    \vspace{-0.3cm}
    \caption{\footnotesize{\textbf{\emph{Performance of scaling compute for single search iteration.}} We use the SiT-XL model, fix the denoising budget to $250$ NFE, and demonstrate the performance differences with respect to the NFEs devoted to a single search iteration.}}
    \label{fig:scale-nfe}
    \vspace{-1.0cm}
\end{wrapfigure}

\textbf{Compute per search iteration.} Within each search iteration, we could adjust the number of denoising steps the model takes. For simplicity, we denote this \textit{NFEs/iter}. Whereas the model performance plateaus quickly when only increasing denoising steps (Figure~\ref{fig:teaser}), we observe that during the search process, adjusting NFEs/iter can reveal distinct compute-optimal regions, as demonstrated in Figure~\ref{fig:scale-nfe}.  Smaller NFEs/iter during search enables efficient convergence, though with a lower final performance. Conversely, larger NFEs/iter result in slower convergence but yield improved performance. Additionally, a diminishing return effect is demonstrated: when NFEs/iter $\geq 50$, further increases in NFEs/iter yield minimal gains despite the additional computational investment. Inspired by this observation, we set the NFEs/iter for each search iteration to $50$ for previous experiments on ImageNet for efficient compute allocation. For experiments in text-to-image setting, since FLUX-1.dev is able to generate high-quality samples with relatively small number of denoising steps, we fix the NFEs/iter to $30$, aligning with the final generation.

\textbf{Compute of final generation.} Despite the freedom in adjusting the denoising steps for the final generation, we always use the optimal setting for the best final sample quality. In ImageNet, we fix $250$ NFEs for the denoising budget, and in text-to-image setting a $30$-step sampler is used, as scaling up further will quickly come to a performance plateau.

\vspace{-0.25cm}
\subsection{Effectiveness of Investing Compute}

We explore the effectiveness of scaling inference-time compute for smaller diffusion models and highlight its efficiency relative to the performance of their larger counterparts without search. For ImageNet tasks, we utilize SiT-B and SiT-L, and for text-to-image tasks, we use the smaller transformer-based model PixArt-$\Sigma$~\citep{chen2024pixart} besides FLUX-1.dev. We report various metrics evaluated on these models under their optimal setups: Zero-Order Search with DINO logits for FID on ImageNet, Random Search with DINO logits for IS on ImageNet, and Random Search with the Verifier Ensemble for text-to-image evaluation on DrawBench. Since models of different sizes incur significantly different costs per forward pass, we use estimated GFLOPs to measure their computational cost instead of NFEs.

\begin{figure}[t]
    \centering
    \begin{minipage}[t]{0.48\linewidth}
        \includegraphics[width=\linewidth]{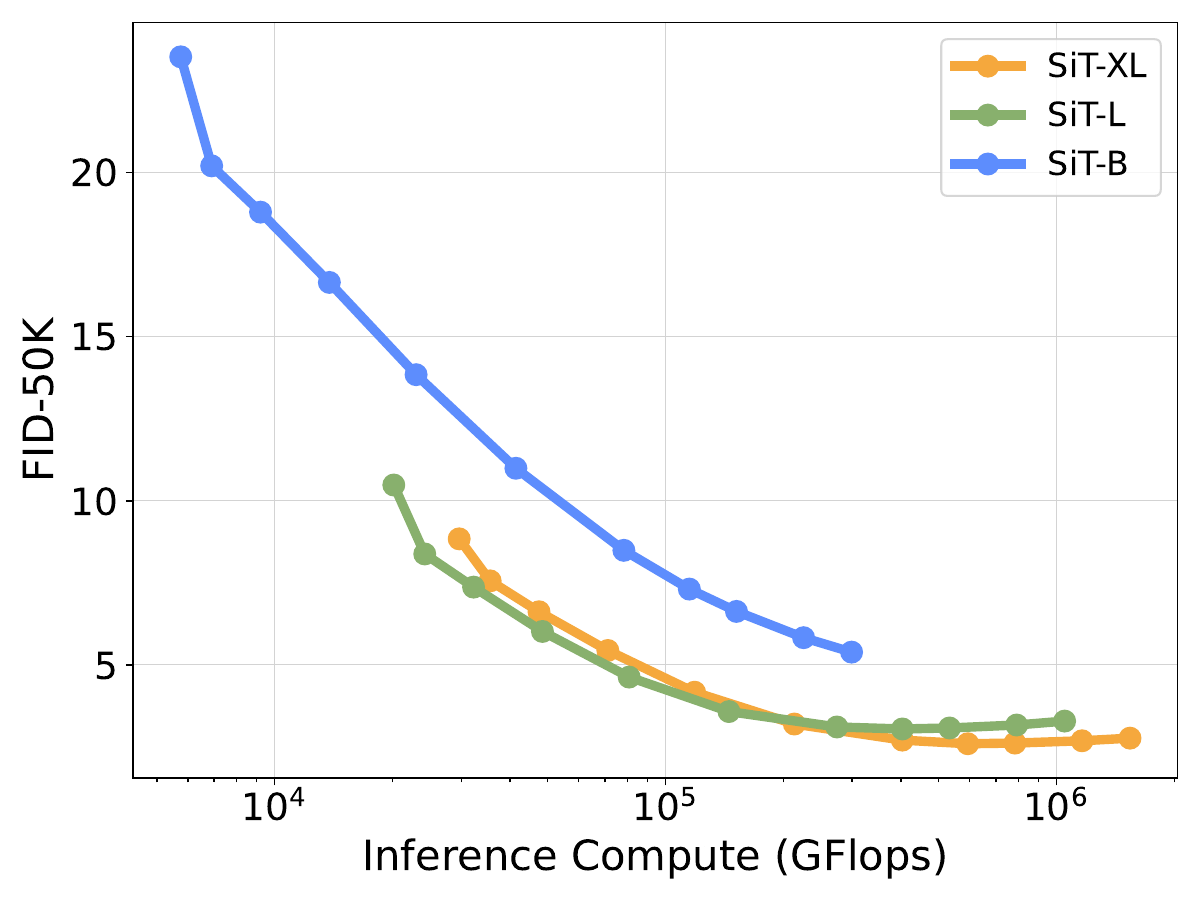}
    \end{minipage}
    \begin{minipage}[t]{0.48\linewidth}
        \includegraphics[width=\linewidth]{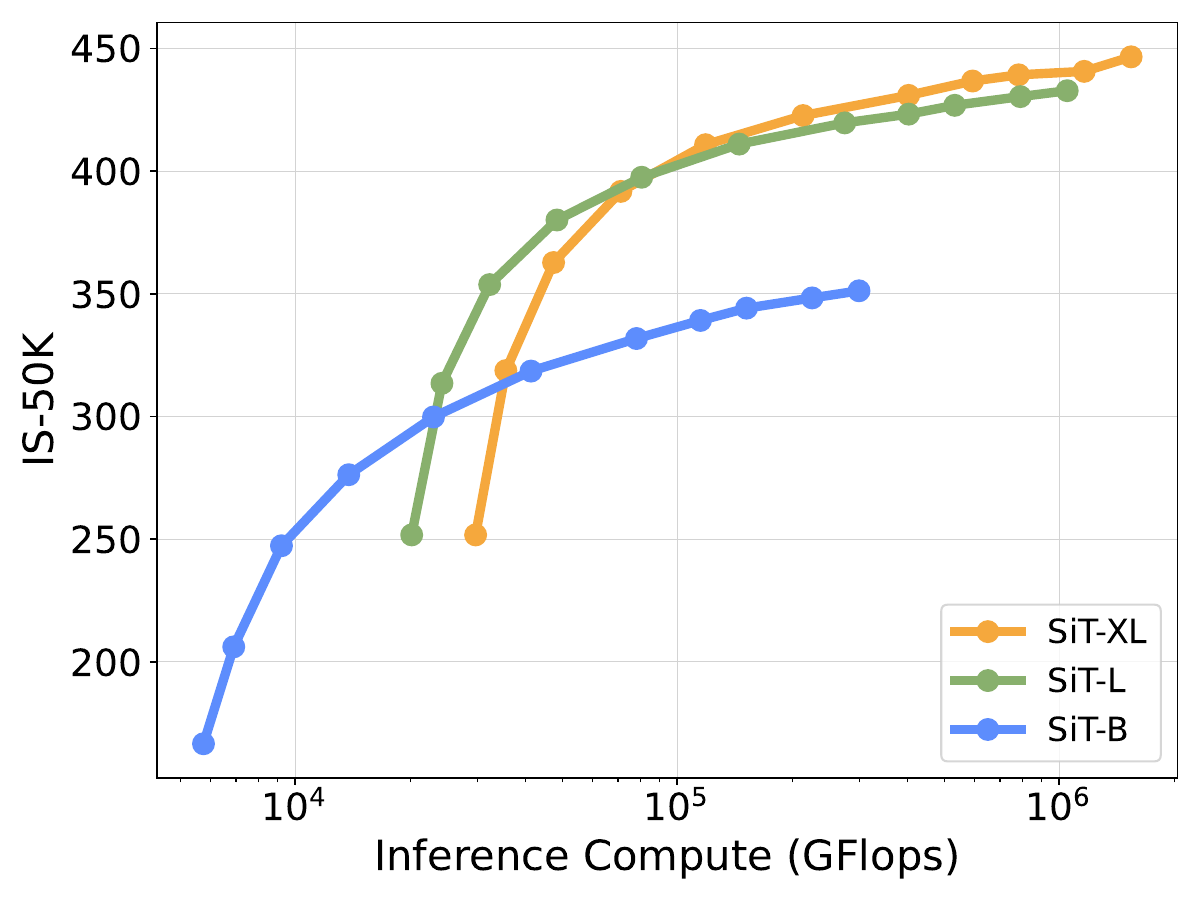}
    \end{minipage}
    
    \caption{\footnotesize{\textbf{\emph{Performance of our search methods across different model sizes (SiT-\{B,L,XL\}) on ImageNet}}. We use the best set up for FID and IS separately. \textbf{Left}: ZO-4 with DINO-LinearHead.; \textbf{Right}: Random Search with DINO-LinearHead.}}
    \label{fig:scale-small-model}
    \vspace{-0.5cm}
\end{figure}

From Figure~\ref{fig:scale-small-model}, we observe that scaling inference-time compute for small models on ImageNet can be highly effective. With a fixed inference compute budget, performing search on small models can outperform larger models without search. For instance, SiT-L demonstrates an advantage over SiT-XL in regions with limited inference compute. However, comparing SiT-B with the other two models reveals that this effectiveness depends on the relatively strong baseline performance of the small models. When a small model's baseline performance lags significantly, the benefits of scaling are limited, resulting in suboptimal outcomes.

\begin{wraptable}{r}{0.5\textwidth}
\centering
\scalebox{0.65}{

\begin{tabular}{lccccc}
\toprule
\multicolumn{1}{c}{\textbf{Model}} & Compute Ratio & Aesthetic & CLIP & ImageReward & LLM Grader\\
\midrule
FLUX & 1 &  5.79 & 0.71 & 0.97 & 84.29\\
\midrule
PixArt-$\Sigma$  & $\sim$0.06 & 5.94 & 0.68 & 0.70 & 84.67\\
 & $\sim$0.09 & 6.03 & 0.71 & 0.97 & 85.62 \\
 & $\sim$2.59 & \textbf{6.20} & \textbf{0.73} & \textbf{1.15} & \textbf{86.95}\\
\bottomrule
\end{tabular}

}
\caption{\footnotesize{\textbf{\emph{Comparison between PixArt-$\Sigma$ when search with Verifier Ensemble and FLUX without search.}}} We use the total compute consumed by FLUX to generate one sample as the standard unit and scale the compute used by PixArt-$\Sigma$ accordingly. These total compute estimates are based on our best approximation and may not be entirely precise.}
\label{tab:scale-pixart}
\vspace{-0.5cm}
\end{wraptable}

These observations extend to the text-conditioned setting, as demonstrated in Table~\ref{tab:scale-pixart}. With just one-tenth of the compute, PixArt-$\Sigma$ outperforms FLUX-1.dev without search, and with roughly double the compute, PixArt-$\Sigma$ surpasses FLUX-1.dev without search by a significant margin. These results have important practical implications: the substantial compute resources invested in training can be offset by a fraction of that compute during generation, enabling access to higher-quality samples more efficiently.

\vspace{-0.25cm}
\section{Related Work}

\textbf{Scaling test-time compute.} Scaling test-time compute is proven to be highly effective on pre-trained LLMs. This presents a completely different axis in LLM's scaling behaviors and inspires many investigations. Recent studies in test-time scaling of LLMs mainly focus on three aspects: (1) better search/planning algorithms~\citep{wei2022chain,gandhi2024stream,su2024dualformer,xie2024monte}; (2) better verifiers~\citep{cobbe2021training,li2024process,wang2023math,lightman2023let}; and (3) scaling law of test-time compute~\citep{wu2024empirical,brown2024large,snell2024scaling}. These works highlight the importance of test-time compute and methods for effectively allocating these compute under a certain budget, orienting the community towards building agents with the ability to reason and self-correct. Inspired by these works, we study the scaling behavior of diffusion models at inference-time, introduce a general search framework over injected noises during sampling, and demonstrate its effectiveness across different benchmarks, aiming to motivate more explorations of inference-time scaling in the diffusion model community.

\textbf{Fine-tuning diffusion models.} To align diffusion models with human preferences, multiple fine-tuning methods have been proposed. \citet{fan023sft} interpret the denoising process as a multi-step decision-making task and use policy gradient algorithms to fine-tune diffusion samplers. \citet{fan2024reinforcement, black2023training} formulate the fine-tuning task as an RL problem, and using policy gradient to maximize the feedback-trained reward. \citet{xu2024imagereward, clark2024directly} further simplifies this task by directly back-propogating the reward function gradient through the full sampling procedure. \citet{wallace2024diffusion} reformulate \textit{Direct Preference Optimization}~\citep{rafailov2023dpo} to derive a differentiable preference objective that accounts for a diffusion model notion of likelihood, and \citet{yang2024using} discard the explicit reward model and directly fine-tune the model on human preference data. Lastly, \citet{carles2024adjoint} casts fine-tuning problem as stochastic optimal control to better align with the tilted distribution of base and reward models. These studies represent substantial advancements in enforcing alignment in diffusion models, ensuring they better adhere to human preferences, ethical considerations, and controlled behaviors.

\textbf{Sample selection and optimization in diffusion models.} The large variation in diffusion's sampling qualities leads to the natural question of how to find good samples during test-time. 
To address this, several works focus on sample selection guided by some pre-defined metrics using the Random Search algorithm. \citet{karthik2023if} and \citet{liu2024correcting} use pre-trained VQA and human preference models to guide the selection, and \citet{liu2024correcting} further update the proposal distribution during selection to better align with the ground truth distribution. Similarly, \citet{na2024diffusion} performs rejection sampling on the updated proposal distribution during intermediate diffusion denoising step. On the other hand, \citet{tang2024realfill} and \citet{samuel2024generating} use a small set of ground truth images as reference and use the similarity between reference and generated images as a guide for selection. 
Yet, these works primarily focus on addressing challenges using very specific verifier and algorithm, while largely overlooking a comprehensive investigation into the biases inherent in different verifiers, the interplay of multiple verifiers and search methods on different tasks, and the relationship between inference-time compute budget and scaling performance.
Some other works~\citep{wallace2023end,ben2024d,novack2024ditto,eyring2024reno,karunratanakul2024optimizing} utilize the gradient of a pre-trained reward model to directly optimize for a better sample. We note, again, that these works focus on relatively small-scaled tasks (in-painting, editing, super-resolution), and the costs of these methods are prohibitive due to the need to back-propagate through the diffusion sampling process.

Recently, several studies\citep{zhou2024goldennoisediffusionmodels, ahn2024noiseworthdiffusionguidance} have proposed approximating the distribution of ``good'' noises using neural networks. These approaches first identify preferable noises $x_T'$ by transforming random noises $x_T \sim \mathcal{N}(0, \mathbf{I})$ through guided DDIM inversion. Subsequently, they train a lightweight predictor on the set of $(x_T, x_T')$ pairs for sampling preferable noises at inference-time. Although these methods shift computational costs from test time to a one-time training phase, they require additional dataset curation and parameter tuning, and can have unsatisfying performance in some application scenarios.

\vspace{-0.5cm}
\section{Conclusion}

In this work, we present a framework for inference-time scaling in diffusion models, demonstrating that scaling compute through search could significantly improve performances across various model sizes and generation tasks, and different inference-time compute budget can lead to varied scaling behavior. Identifying verifiers and algorithms as two crucial design axes in our search framework, we show that optimal configurations vary by task, with no universal solution. Additionally, our investigation into the alignment between different verifiers and generation tasks uncovers their inherent biases, highlighting the need for more carefully designed verifiers to align with specific vision generation tasks.

\vspace{-0.2cm}
\section*{Acknowledgements}
\vspace{-0.2cm}
We thank Ziyu Wan, Jack Lu, Boyang Zheng, Oliver Wang, Jason Baldridge and Sayak Paul for their insightful discussions.

\bibliographystyle{abbrvnat}
\nobibliography*
\bibliography{googledeepmind-test}

\newpage 

\appendix 
\part*{Appendices}

\section{Experiment Settings}
\label{app:sec:exp-setting}

We present our experimental settings below.

\vspace{-0.25cm}
\subsection{Training Settings}

Most models used in our work are pre-trained: in ImageNet, we use the pre-trained SiT-XL model; under Text-to-Image setting, we use the publicly released weights of FLUX.1-dev and PixArt-$\Sigma$ from the \texttt{diffusers} library~\citep{von-platen-etal-2022-diffusers}. In Section~\ref{sec:dimension}, the reported SiT-B and SiT-L are self-trained following the identical architectures and training configurations from~\citep{ma2024sit}. The final numbers included in Figure~\ref{fig:scale-small-model} are from models trained at 800K iterations.

\vspace{-0.25cm}
\subsection{Sampling Settings}

We summarize the sampling settings in our work below.

\begin{table}[h]
    \centering
    \scalebox{0.9}{
    \begin{tabular}{cccc}
    \toprule
    \textit{Configs} & \textit{Class-conditioned} & \multicolumn{2}{c}{\textit{Text-conditioned}} \\
    
    & SiT-XL & FLUX.1-dev & PixArt-$\Sigma$ \\
    \midrule
    ODE solver & 2$^\text{nd}$ order Heun &  Euler & DDIM \\
    NFEs/iter & $50^\dagger$ & $30$ & $30$ \\
    final denoising steps & $250$ & $30$ & $30$ \\
    guidance scale & $1.0^\ddagger$ & $3.5$ & $4.5$ \\
    resolution & $256$ & $1024$ & $1024$\\ 
    \bottomrule
    \end{tabular}
    }
    \caption{\footnotesize{\textbf{\emph{Default sampling settings for Class-conditioned and Text-conditioned generation.}} $\dagger$ In Figure~\ref{fig:scale-nfe} we report numbers with different NFEs/iter; $\ddagger$ In Figure~\ref{fig:search-classifier} we report results with different guidance scales.}}
    \label{tab:sample-setting}
    \vspace{-0.5cm}
\end{table}

\subsection{Search Settings}

\textbf{Random Search.} During search, we randomly sample a set of i.i.d Gaussian noises $|S|$ as the candidates for each conditioning, generate samples from them with the ODE solver, and select the one with the highest score output by the verifiers as the noise used for final generation. We take the size of $S$ as the primary scaling axis and explore $|S| = 2^k$ for $k \in \{1,2,3,4,5,6,7,8\}$ in our experiments.

\begin{figure}[t]
    \centering
    \begin{minipage}{0.48\linewidth}
    \includegraphics[width=0.48\linewidth]{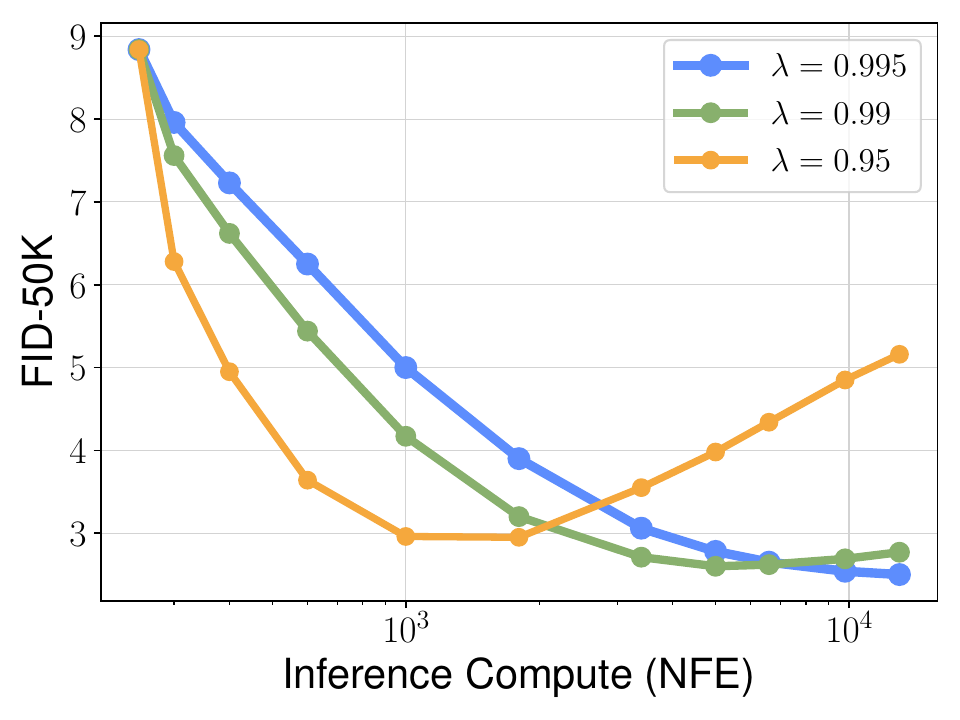}
    \includegraphics[width=0.48\linewidth]{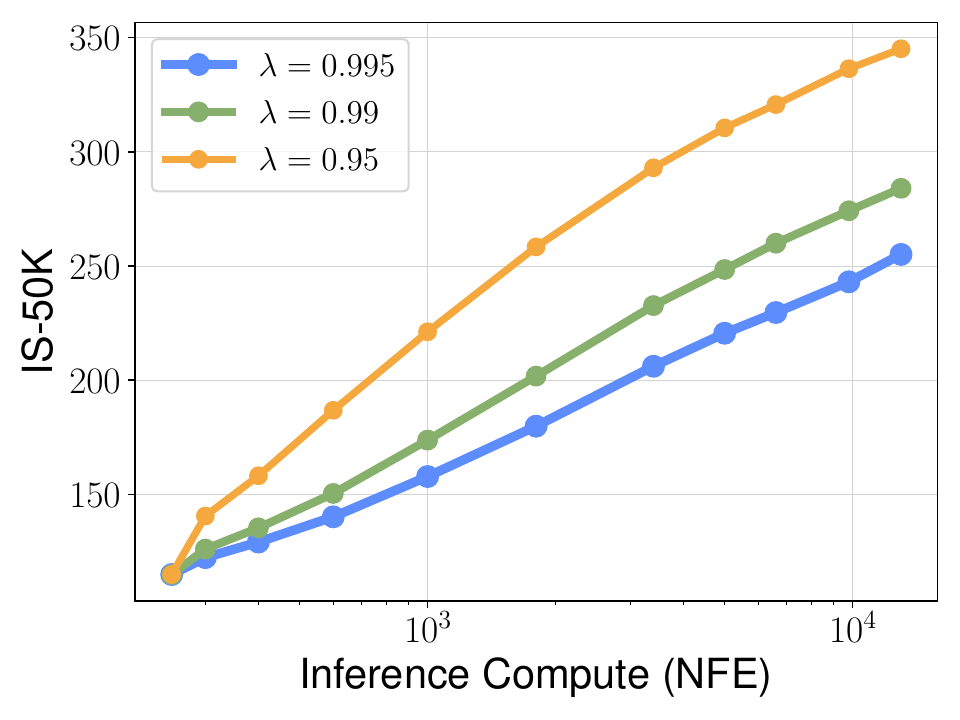}
    \end{minipage}
    \begin{minipage}{0.48\linewidth}
    \includegraphics[width=0.48\linewidth]{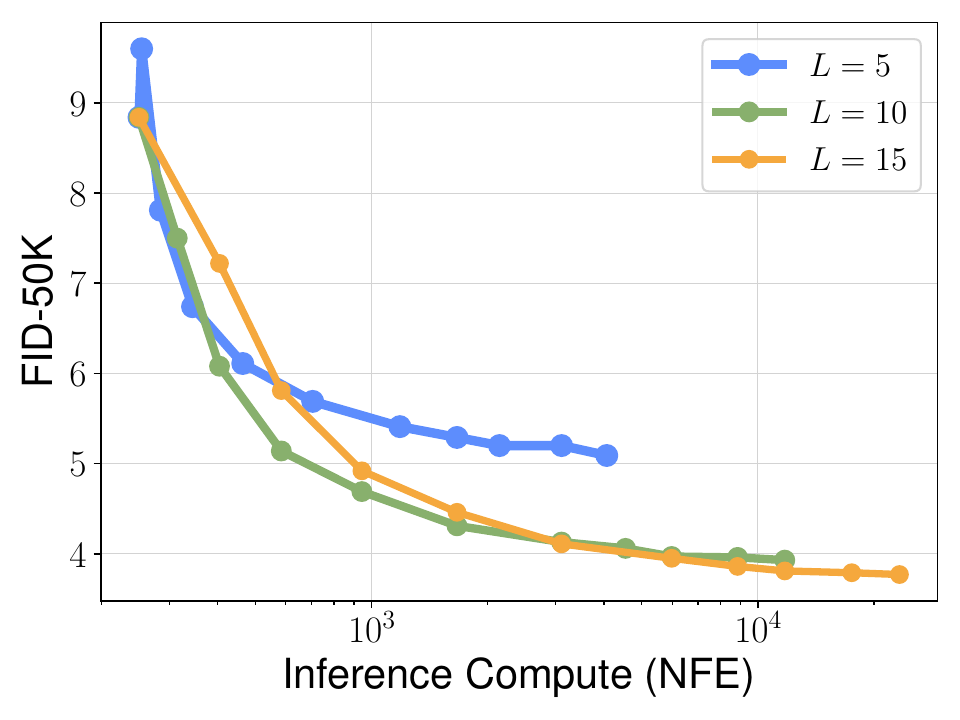}
    \includegraphics[width=0.48\linewidth]{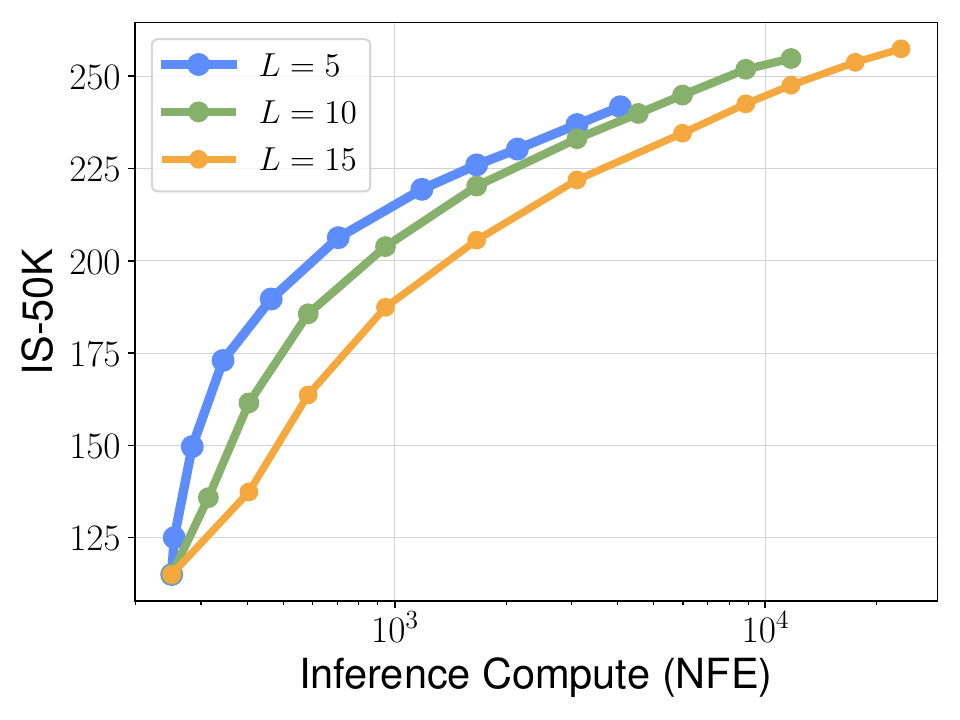}
    \end{minipage}
    \caption{\footnotesize{\textbf{\emph{Performance of tuning additional parameters for algorithms}. } \textbf{Left}: Zero-Order Search with step
sizes $\lambda$; \textbf{Right}: Search Over Paths with lengths $L$. We use SiT-XL and fix the verifier to be the classification logits from DINO.}}
    \label{fig:algo-ablation}
    \vspace{-0.5cm}
\end{figure}

\textbf{Zero-Order Search.} There are three tunable parameters in Zero-Order Search: search iterations $K$, number of neighbors $N$, and step size $\lambda$. As $K$ is the most scalable, we fix it as the primary scaling dimension when studying the behavior of Zero-Order Search. In Figure~\ref{fig:search-algos} we additionally investigate the performance of Zero-Order Search when tuning $N$, as it provides a secondary axis in scaling compute. In Figure~\ref{fig:algo-ablation}, we demonstrate the effect of tuning step size $\lambda$. We fix $N = 2$ and explore the performance of modifying the values of $\lambda$ with respect to different values of $K$.

Expectedly, when $\lambda$ is small, Zero-Order Search has slightly worse performance and lower compute efficiency; when $\lambda$ is large, Zero-Order Search suffers from overfitting - the selected set of noises fits too close to the high scoring area of the verifier, leading to loss of diversity. As a result, while it has the best Inception Score among the three, its FID starts to increase once the compute is scaled over $10^3$ NFEs. We provide further analysis of this observation in Section~\ref{app:sec:fid-loss-diversity}.

\textbf{Search Over Paths.} We summarize the hyperparameters for Search Over Paths below

\begin{table}[h]
    \centering
    \scalebox{0.75}{
    \begin{tabular}{lcl}
    \toprule
    \multicolumn{2}{c}{\textit{Hyperparameter}} & \multicolumn{1}{c}{\textit{Description}} \\
    \midrule
    initial paths  & $N$ & The number of paths to start the search with. \\
    paths width & $M$ & The number of noises to sample within each path. \\
    search start & $\sigma$ & The time to start search. \\
    backward stepsize & $\Delta b$ & The length of time interval to run ODE solver. \\
    forward stepsize & $\Delta f$ & The length of time interval to run noising process. \\
    paths length & $L$ & The NFEs devoted in each backward step. \\
    \bottomrule
    \end{tabular}
    }
    \caption{\footnotesize{\textbf{\emph{Hyperparameters for Search Over Paths.}}}}
    \label{tab:paths-setting}
\end{table}

For $N > 1$, we start search with $N$ i.i.d samples Gaussian noises and obtain $N$ noisy samples $\{x_\sigma^i\}$. For each $x_\sigma$, we then formulate its sampling paths and search over them. Once the search terminates, $N$ different $\hat{x}_0$ are left, and we run them through the Best-of-N selection to obtain the best one.

In our experiments, we set $M$ and $N$ to be our primary and secondary scaling axis, respectively, as shown in Figure~\ref{fig:search-algos}. We further explore the behavior of tuning the paths length in Figure~\ref{fig:algo-ablation}, where we see that scaling up the paths length can be beneficial to FID but have marginal effect on Inception Score. This supports our claim that the search settings need to be specifically tuned to different application scenarios.

We fix other hyperparameters: $\sigma = 0.11$, $\Delta b = 0.81$, $\Delta f = 0.78$, inspired by the setting in~\citep{xu2023restart}.

\vspace{-0.25cm}
\subsection{Verifier Settings}

\textbf{ImageNet.} We consider a total of four verifiers for search on ImageNet. We list the settings below: \begin{itemize}
    \item \textbf{FID}: we denote the ground truth feature statistics for ImageNet training set $\mu_{\text{ref}}$ and $\Sigma_{\text{ref}}$. During search, we select the first $1024$ samples randomly and use them to initialize a running mean and covariance $\hat{\mu}$ and $\hat{\Sigma}$. In the following search iterations, for each candidate batch $b_i$ a staged mean $\hat{\mu}_i$ and covariance $\hat{\Sigma}_i$ are calculated with the batch information and the previous running mean and variance. A corresponding $\text{FID}_i$ will be obtained between $\hat{\mu}_i$, $\hat{\Sigma}_i$ and $\mu_{\text{ref}}$, $\Sigma_{\text{ref}}$, which is then used as the verifier score. Eventually, $b = \arg\min_i \text{FID}_i$ will be selected, and $\mu_{\text{ref}}$ and $\Sigma_{\text{ref}}$ are then updated accordingly. Such iteration is repeated until we reach $50000$ samples.
    
    \item \textbf{IS}: The class confidence probability output by the InceptionV3 model is taken as the verifier score.
    
    \item \textbf{CLIP}: For logits, we use the zero-shot classifier weight $W$ generated by prompt engineering specified in~\citet{radford2021learning} and take the cosine-similarity between the corresponding class entry in $W$ and the image feature to be the verifier score. For the self-supervised version, we directly extract the image feature.
    
    \item \textbf{DINO}: For logits, we use the pre-trained linear classification head provided in~\citet{oquab2023dinov2} as the verifier. Specifically, we concatenate the \texttt{cls} tokens from the last four layers along with the average pooling of the feature from the last layer to formulate the input for the linear head. For the self-supervised version, we directly take the \texttt{cls} token from the last layer.
\end{itemize}

\textbf{Text-to-Image.} We consider a total of four verifiers for search in Text-to-Image setting: \begin{itemize}
    \item \textbf{Aesthetic}: we take the aesthetic predictor pre-trained on subset of LAION-5B. It consists of a single MLP without any non-linearity and takes the image feature from a pre-trained CLIP-L model as input. The output is on a scale of $0-10$ rating the images' aesthetic quality.
    \item \textbf{CLIPScore}: we take the pre-trained CLIP-L model and measure the cosine similarity between visual and text features. Following~\cite{hessel2021clipscore}, each text prompt is additionally prefixed with \texttt{'A photo depicts'}, and the final score is rescaled by \texttt{2.5 * max(cos\_sim, 0)}.
    \item \textbf{ImageReward}: we take the pre-trained model for approximating human preference from~\cite{xu2024imagereward} and use the identical evaluation setting. 
    \item \textbf{Verifier Ensemble}: We separately run candidates through the above three verifiers, rank the scores output by each, and use the unweighted average rankings as the final score for the Verifier Ensemble.
\end{itemize}

\vspace{-0.25cm}
\subsection{Evaluation Setting}

\textbf{ImageNet.} Following standard practice, we calculate FID and Inception Score using $50000$ synthesized samples. We use randomly generated conditions and a global batch size of $256$ for all evaluations. We extracted the ImageNet statistics and calculated FID and IS following~\citet{karras2024analyzing}.

\textbf{DrawBench.} We search for one noise per prompt for generating the sample. For evaluators other than the \textit{LLM Grader}, we simply input the synthesized samples into the pre-trained evaluator models and report the averaged scores over the $200$ prompts.

For \textit{LLM Grader}, we prompt the Gemini-1.5 flash model to assess synthesized images from five different perspectives: Accuracy to Prompt, Originality, Visual Quality, Internal Consistency, and Emotional Resonance. Each perspective is rated from $0$ to $100$, and the averaged overall score is used as the final metric. We include the break-down scores in Table~\ref{tab:llm-break-down}, and in Figure~\ref{fig:gemini-prompt} we present the detailed prompt. We observe that search can be beneficial to each scoring category of the \textit{LLM Grader}.

\textbf{T2I-CompBench.} For each prompt we search for two noises and generate two samples. During evaluation, the samples are splitted into six categories: \texttt{color},  \texttt{shape}, \texttt{texture}, \texttt{spatial}, \texttt{numeracy}, and \texttt{complex}. Following~\citet{huang2023t2i}, we use the BLIP-VQA model~\citep{li2022blip} for evaluation in \texttt{color},  \texttt{shape}, and \texttt{texture}, the UniDet model~\citep{zhou2022simple} for \texttt{spatial} and \texttt{numeracy}, and a weighted averaged scores from BLIP VQA, UniDet, and CLIP for evaluating the \texttt{complex} category.

\begin{table*}[t!]
    \centering
    \scalebox{0.9}{
    \begin{tabular}{lr|cccccc}
    \toprule
    \multicolumn{2}{c}{Model} & Accuracy$\uparrow$  & Originality$\uparrow$ & Visual$\uparrow$ & Consistency$\uparrow$ & Emotional$\uparrow$ & Overall$\uparrow$\\
    \midrule
    \textbf{FLUX.1-dev} &  &  89.35 & 67.58 & 93.00 & 97.04 & 73.99 & 84.29\\
         & + 4 search iters & 91.33 & 68.49 & 93.42 & 96.99 & 75.31 & 85.17\\
         & + 16 search iters & 91.95 & 71.52 &\textbf{ 93.76} & \textbf{97.24} & 76.30 & 86.42\\
         & + 64 search iters & \textbf{93.83} & \textbf{75.38} & 93.57 & 97.04 & \textbf{79.34} & \textbf{88.08}\\
    \midrule
    \textbf{PixArt-$\Sigma$} & & 84.60 & 73.29 & 91.91 & 95.80 & 76.34 & 84.67\\
        & + 4 search iters & 87.88 & 74.03 & 91.92 & 96.29 & 77.32 & 85.62\\
         & + 16 search iters & 88.15 & 75.39 & 91.72 & 96.04 & 79.17 & 86.27 \\
         & + 64 search iters & \textbf{89.30} & \textbf{77.79} & \textbf{92.46} & \textbf{96.68} & \textbf{80.43} & \textbf{87.55}\\
    \bottomrule
    \end{tabular}
    }
    \caption{\footnotesize{\textbf{\emph{Break-down scores of LLM Grader for FLUX.1-dev and PixArt-$\Sigma$.}} The evaluation is done on DrawBench with random search and verifier ensemble.}}
    \label{tab:llm-break-down}
\end{table*}

\vspace{-0.5cm}
\section{Verifier Hacking Leads to Degeneracy in Evaluation Metrics}
\label{app:sec:fid-loss-diversity}

Many prior works~\citep{gao2023scaling, black2023training, kim2024confidence} noticed the overoptimization issue when finetuning diffusion models using pre-trained reward models, that excessively optimizing against a reward model will lead to degeneracy in other evaluation metrics. We have similar observations when we excessively search against a verifier and quickly overfit to its bias.

\begin{figure}[t!]{}
    \centering
    \begin{minipage}{0.48\linewidth}
    \includegraphics[width=0.48\linewidth]{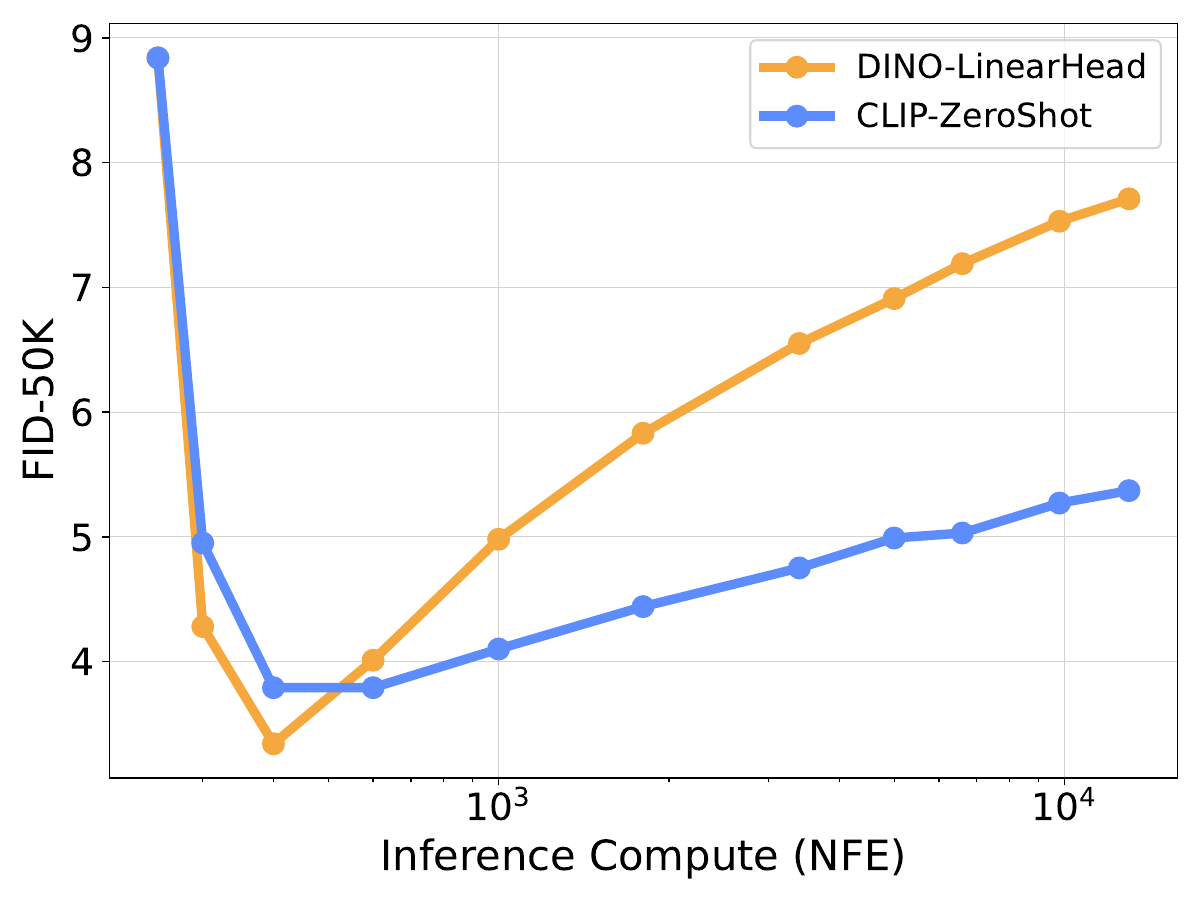}
    \includegraphics[width=0.48\linewidth]{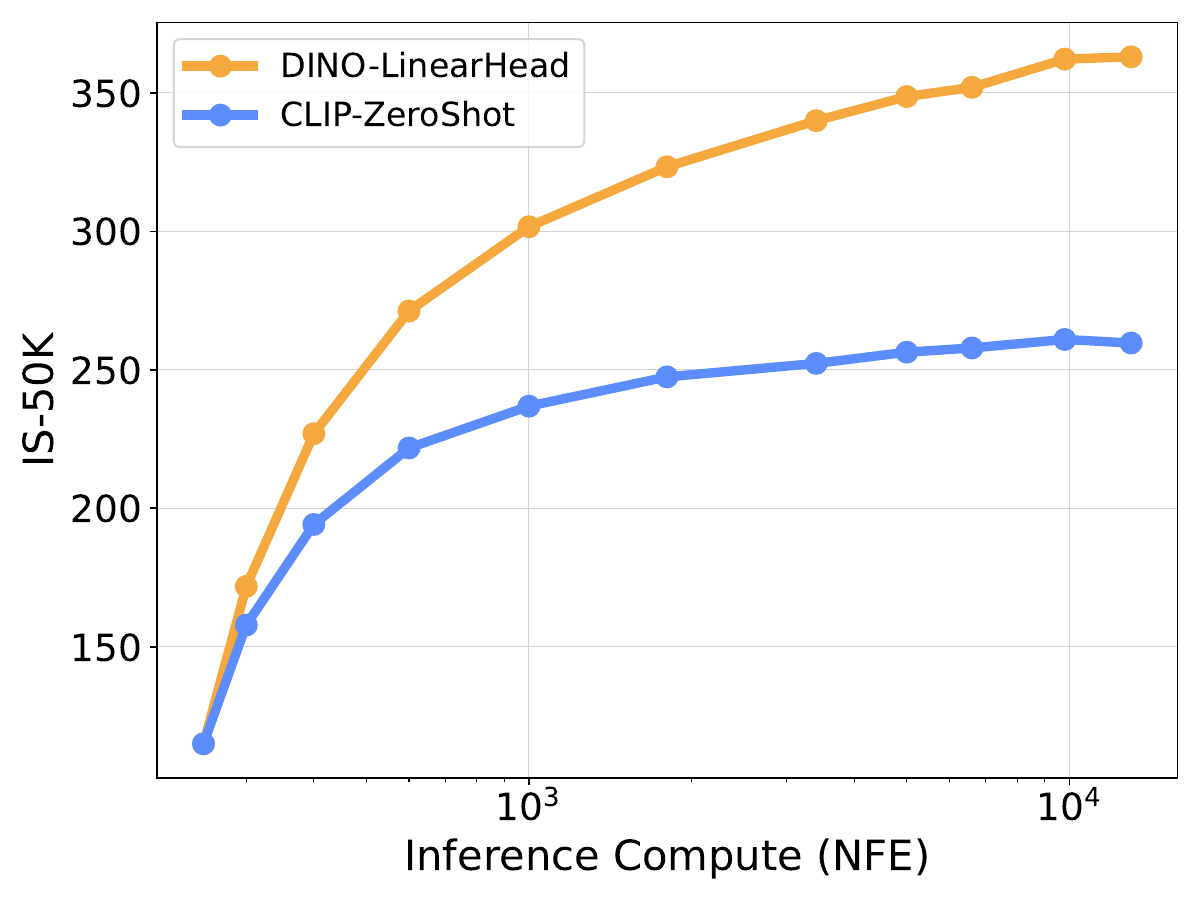}
    \end{minipage}
    \begin{minipage}{0.48\linewidth}
    \includegraphics[width=0.48\linewidth]{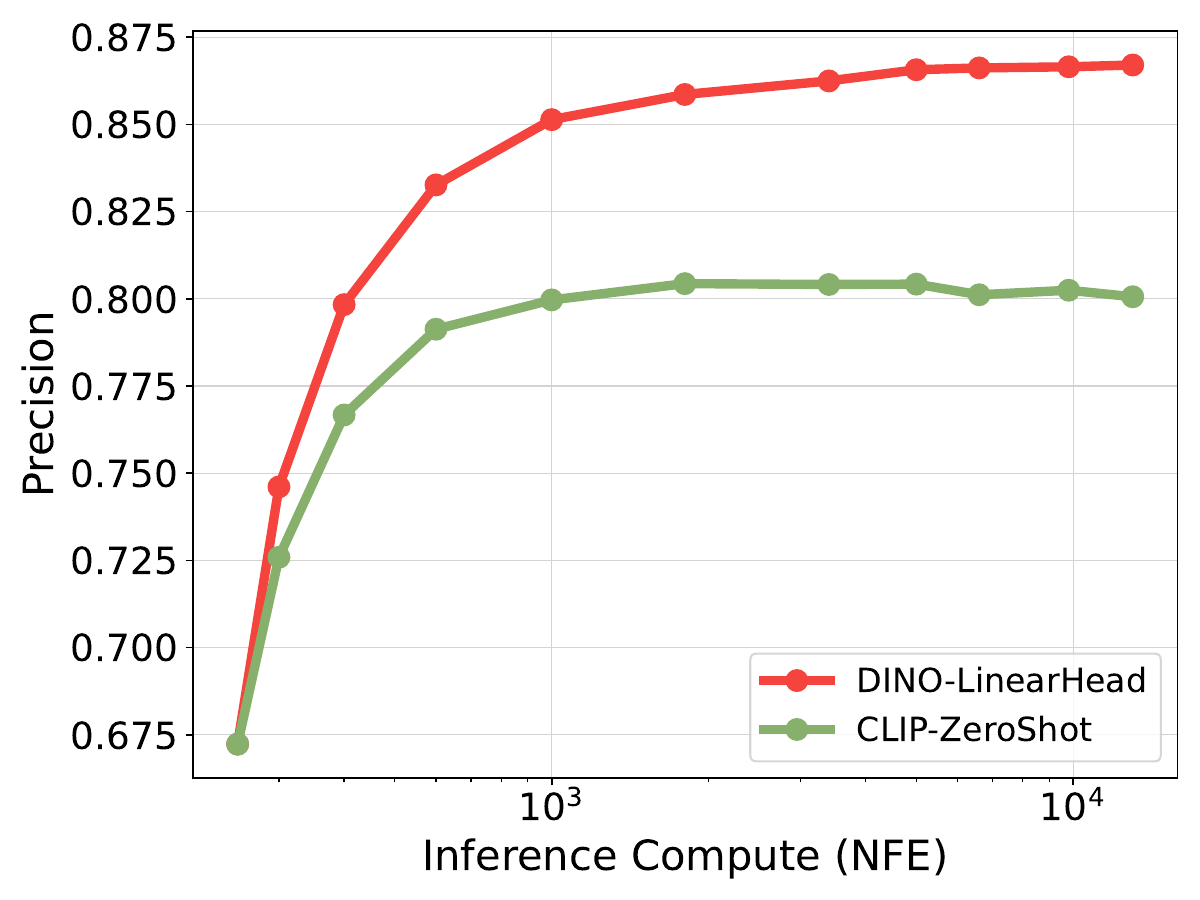}
    \includegraphics[width=0.48\linewidth]{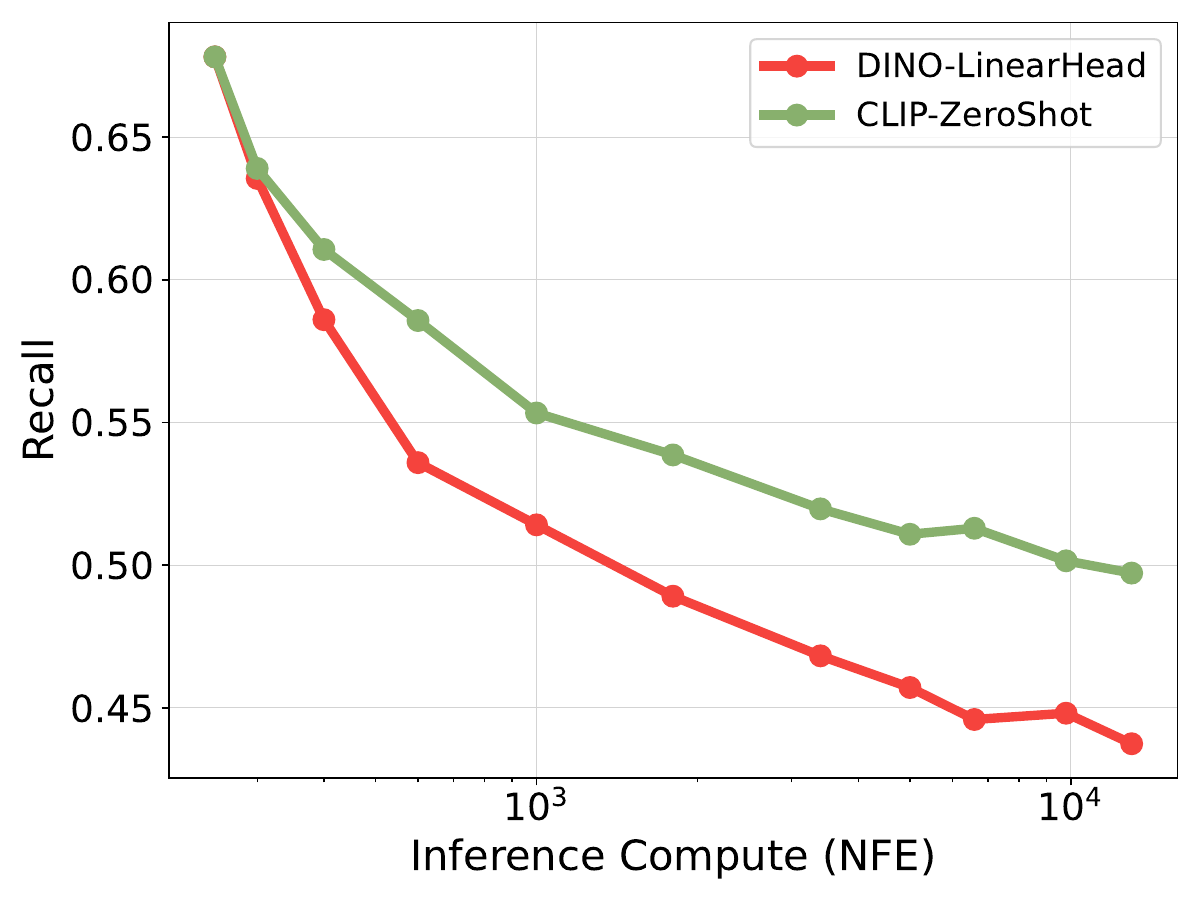}
    \end{minipage}
    \caption{\footnotesize{\textbf{\emph{Performance of Random Search on ImageNet against DINO and CLIP classification logits.}}} We use random search on the SiT-XL model and report FID, IS, Precision, and Recall.}
    \label{fig:random-loss-diversity}
\end{figure}

When search on ImageNet against the DINO or CLIP classification logits, we notice the sudden increasing in FID score once pass a certain search iteration numbers despite the constantly improving Inception Score, as shown in Figure~\ref{fig:random-loss-diversity}. To investigate this issue, we calculate the Precision and Recall~\citep{kynkaanniemi2019improved} and plot them in Figure~\ref{fig:random-loss-diversity}. We see that while Precision increases with search iterations, demonstrating the consistent improvement in sample quality, Recall decreases with search iterations, implying the loss of diversity of the sample set.

We credit this to the DINO and CLIP classification verifiers. When searching against these verifiers, we only operate on a per-noise basis - select the one noise whose corresponding sample has the highest classification score. Therefore, when the search iterations increase, our final set of the selected noises will get closer to the high scoring regions of the classification verifiers. This have two consequences: 1) the selected noises overfit to the verifiers and degenerate other metrics; 2) the selected noises cluster around the high scoring regions and disregard the overall variance of the final set. We deem the latter to be more impactful on the evaluated FID, since FID is known to take great account of the diversity of the generated samples. The Zero-Order Search and Search over Paths we proposed in Section~\ref{sec:algorithm} alleviate this issue to some extent by searching in the local neighborhood of the Gaussian noise $\mathbf{n}$ sampled at the beginning or at the intermediate sampling steps. However, if we expand the neighborhood range for Zero-Order Search as shown in Figure~\ref{fig:algo-ablation}, it will suffer from the diversity issue as well.

A more fundamental solution would be to use the verifiers operating on a population basis and taking into account of the global structure of the set of selected noises. From the trivial example in Figure~\ref{fig:BoN-FID-IS}, we see that such verifiers could be effective. We leave further exploration to future works.

Consequence (1) is better demonstrated in Figure~\ref{fig:flux-metrics}. We see that over-search against Aesthetic Score will lead to degeneracy in CLIPScore, and vice versa.

\vspace{-0.5cm}
\section{Zero-Order and First-Order Search}
\label{app:sec:first-order}

Since many verifiers are differentiable, we also investigate First-Order Search on ImageNet guided by the gradient of verifiers. Specifically: \begin{enumerate}
    \item we initialize the noise prior with a randomly sampled Gaussian vector $\mathbf{n}$
    \item run $\mathbf{n}$ through the diffusion ODE solver to obtain the sample and its corresponding score output by verifier $\mathcal{V}$
    \item backpropogate through the verifier and the ODE solver to calculate $\nabla_\mathbf{n} \mathcal{V}(\mathbf{n})$
    \item update $\mathbf{n}$ via gradient descent: $\mathbf{n}' = \mathbf{n} - \eta \nabla_\mathbf{n} \mathcal{V}(\mathbf{n})$, and repeat step 2-4.
\end{enumerate}

Due to the iterative nature of diffusion sampling process, step 2 will incur prohibitive memory cost if naively backpropogating through the ODE solver. To alleviate this issue, we perform gradient checkpointing~\citep{chen2016training} on each ODE integration step following~\citep{wallace2023end, ben2024d, novack2024ditto}. This discards the intermediate activation values and re-calculate them using one extra model forward call during backpropogation, thus greatly reducing space complexity at the cost of slightly increased execution time.

We also note that performing naive gradient descent in step 4 might push the updated $\mathbf{n}'$ outside the Gaussian manifold, resulting in training and sampling inconsistency. To resolve this, we simply rescale $\mathbf{n}'$ so that its norm is consistent with the norm of i.i.d Gaussian vectors$^6$.

\begin{figure}[t]
    \centering
    \begin{minipage}{0.48\linewidth}
    \includegraphics[width=0.9\linewidth]{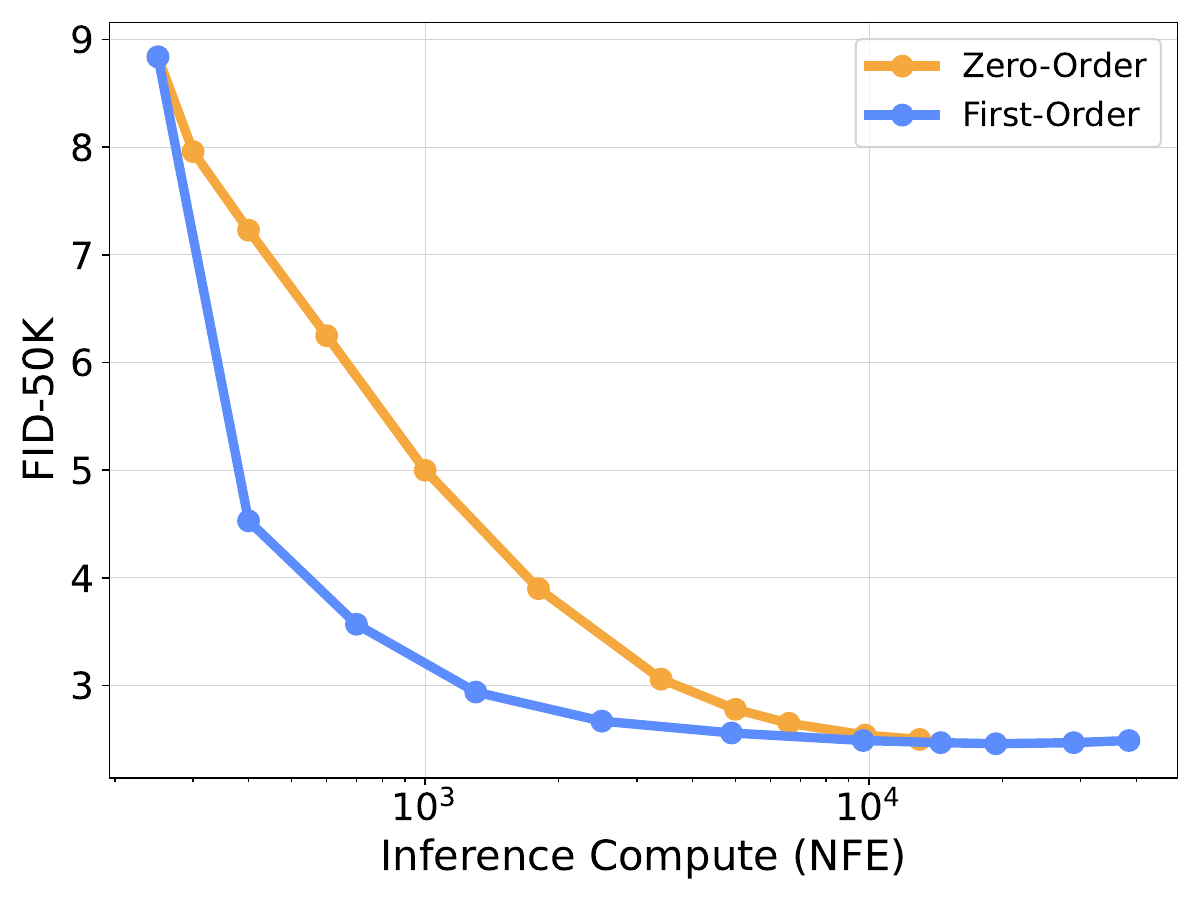}
    \end{minipage}
    \begin{minipage}{0.48\linewidth}
    \includegraphics[width=0.9\linewidth]{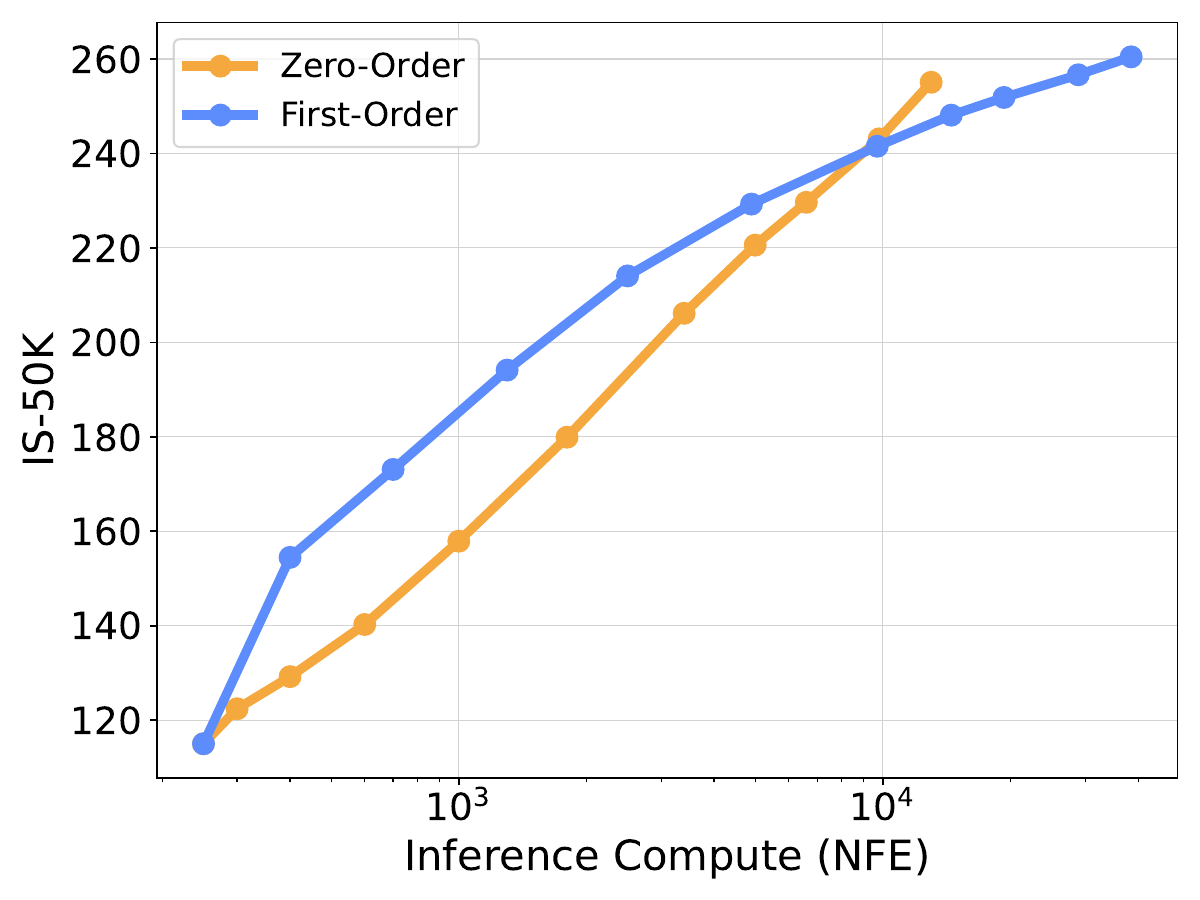}
    \end{minipage}
    \caption{\footnotesize{\textbf{\emph{Comparison between Zero-Order and First-Order Search.}} We use the SiT-XL model and fix the verifier to be the DINO-LinearHead. The Inference Compute is aligned via the rough estimation of cost(backward) $\sim$ 3$\times$cost(forward).}}
    \label{fig:zero-first-comp}
    \vspace{-0.5cm}
\end{figure}

In Figure~\ref{fig:zero-first-comp} we include the comparison between Zero-Order Search and First-Order Search. We fix the learning rate to be $\eta = 0.01$ for First-Order Search to roughly match the step size of Zero-Order Search with $\lambda = 0.995$. At a best estimation we attribute the overhead of gradient checkpointing as twice the number of model forward calls, making each iteration $3\times$ costly than without backpropogation.

With inference compute roughly aligned, although First-Order Search shows faster convergence speed over Zero-Order, we see that it does not demonstrate a significant margin when continuously scaling up compute, despite its higher memory cost$^7$ and worse scalability on large models. However, by its gradient-guided nature, First-Order Search can be advantageous in tasks with more fine-grained objectives, such as image editing, inpainting, and solving inverse problems~\citep{wallace2023end, novack2024ditto, ben2024d, karunratanakul2024optimizing}.

\fancypagestyle{footnote6}{
\fancyfoot[L]{\footerfont \itshape 
$^6$In $\mathbb{R}^d$, the norm of isotropic Gaussian vectors is distributed according to the chi-squared distribution with $d$ degrees of freedom. \\
$^7$Gradient checkpointing still requires $O(\sqrt{n})$ space complexity~\citep{chen2016training}, with $n$ being the number of layers inside the model.}
}
\thispagestyle{footnote6}

\vspace{-0.5cm}
\section{Self-Supervised Verifiers have Marginal Effect in Text-to-Image Setting}
\label{app:sec:t2i-ssl}

Following Section~\ref{sec:verifiers}, we investigate the performance of self-supervised verifiers in text-to-image setting. Apart from DINO and CLIP, we additionally incorporate SigLIP~\citep{zhai2023sigmoid} as an extension to CLIP. Different from ImageNet where the self-supervised verifiers are good surrogate for classification verifiers, on DrawBench they do not demonstrate the expected performance, as shown in Table~\ref{tab:ssl-t2i}.

\begin{wrapfigure}{l}{0.5\textwidth}
    \centering
    \begin{minipage}{0.32\linewidth}
    \includegraphics[width=\linewidth]{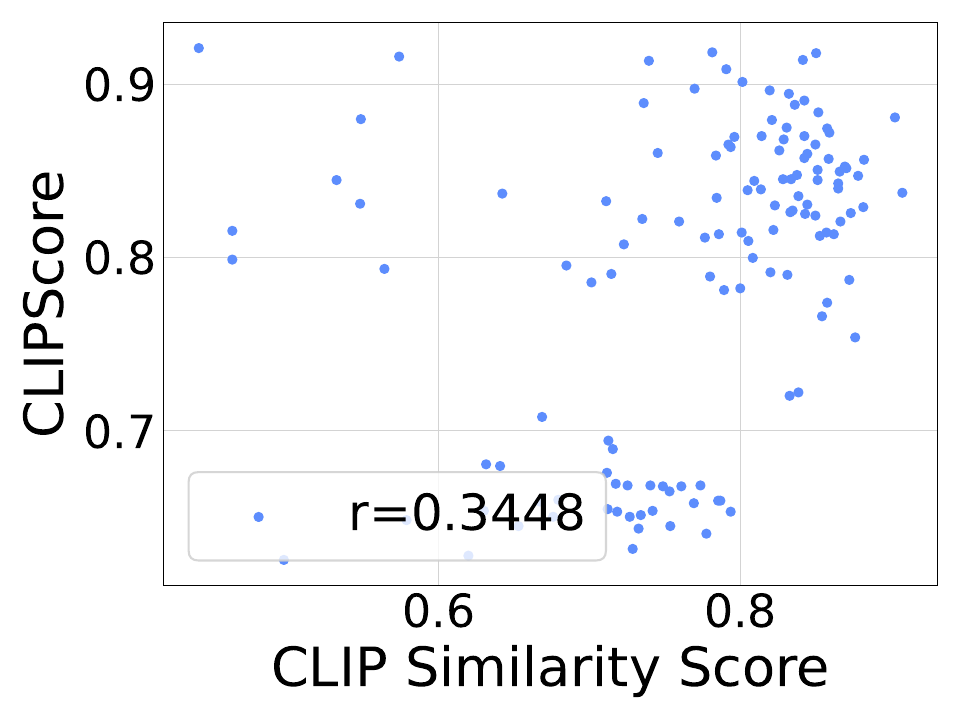}
    \end{minipage}
    \begin{minipage}{0.32\linewidth}
    \includegraphics[width=\linewidth]{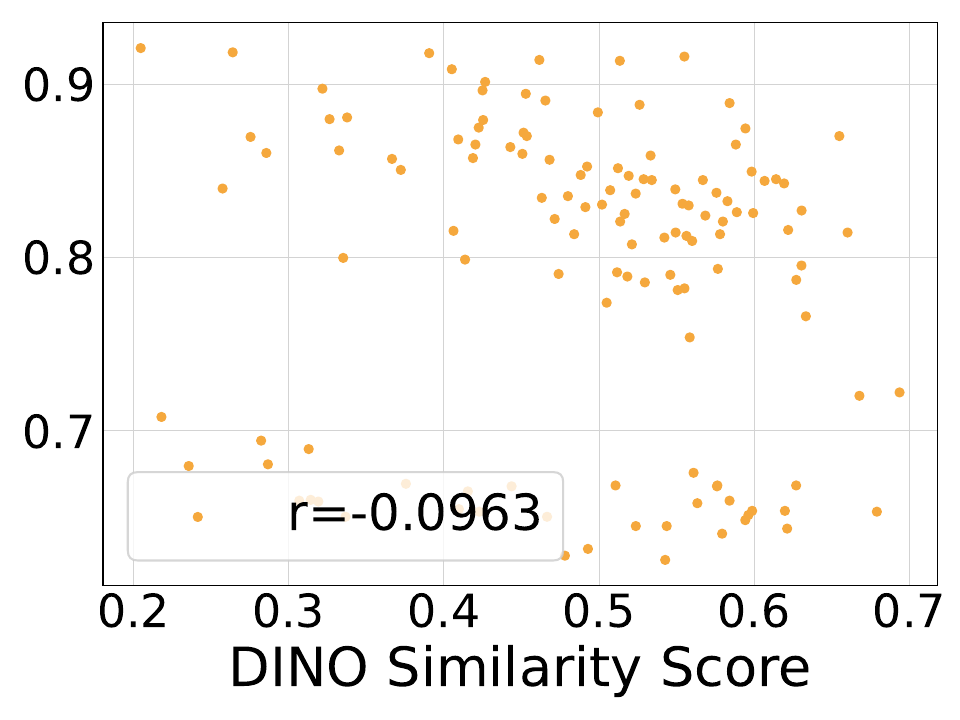}
    \end{minipage}
    \begin{minipage}{0.32\linewidth}
    \includegraphics[width=\linewidth]{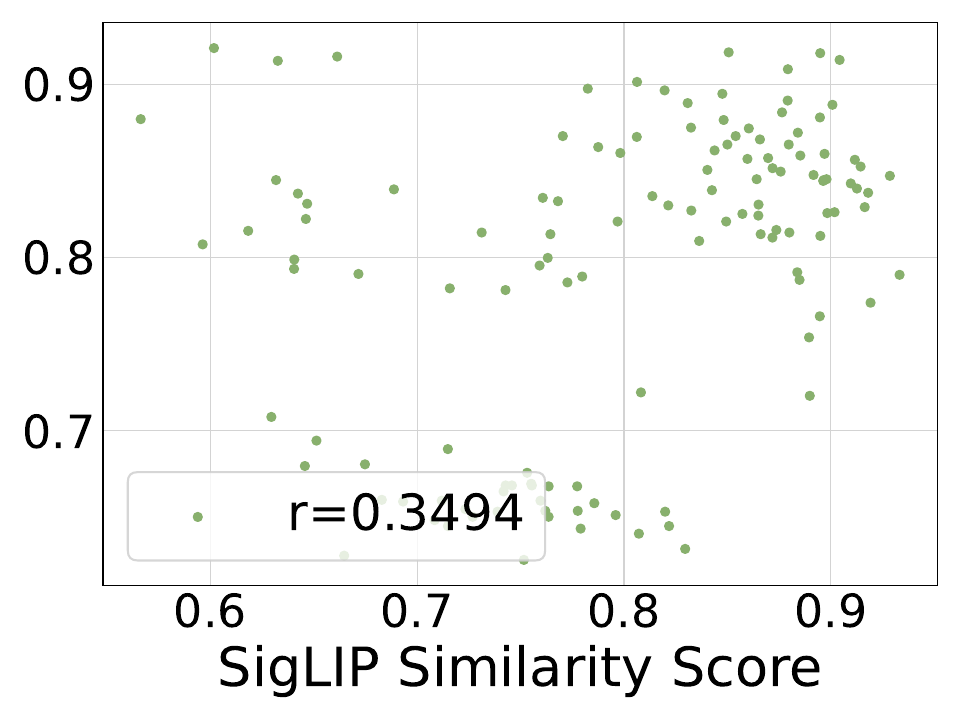}
    \end{minipage}
    \caption{\footnotesize{\textbf{From Left to Right}: Correlation of CLIP, DINO, and SigLIP feature similarity score with CLIPScore. All points are generated from FLUX.1-dev.}}
    \label{fig:t2i-ssl-cor}
    \vspace{-0.5cm}
\end{wrapfigure}

From Figure~\ref{fig:t2i-ssl-cor}, we observe much weaker metric correlations comparing with self-supervised verifiers in ImageNet. We credit this observation to the following: 

1) evaluation metrics in text-to-image settings usually focus on more nuanced perspectives: visual-text alignment, composition correctness, human preferences, etc. Even the aesthetic predictor has its bias - it prefers stylized images over others~\citep{clark2024directly}. On the other hand, self-supervised verifiers essentially select the samples with smallest trajectory curvature in the feature space, which implies a stabler sampling process and thus potentially higher sample quality. Yet, by the subtle and holistic nature of evaluation in text-to-image settings~\citep{lee2024holistic}, such "higher sample quality" may not align with the specific perspectives each metric focuses on. For example, under the same text prompt, an image with high visual quality but misaligned content might not be preferred over an image with slightly degraded visual quality but richer and more aligned visual content.

2) The rich conditionings and extensive fine-tuning in text-to-image models on large scale datasets might lead to different sampling dynamics comparing to the small class-conditioned models trained on ImageNet. This may lead to failure of the self-supervised verifier themselves, as the low trajectory curvature measured in feature space might no longer be indicative of the sample quality.

This also calls for the design of task specific verifiers. From the self-supervised verifiers across class-conditioned and text-conditioned tasks, we see that the effectiveness of verifiers can be highly task-dependent. Therefore, to conduct search that's better aligned with desired objectives, we deem it necessary to have verifiers designed specifically for each task; during search, it's also very important to avoid hacking to the specific bias of each verifier. We have proposed some simple methods in our work, and we leave further explorations of this problem to future works.

\begin{table}[t!]
    \centering
    \scalebox{0.95}{
    \begin{tabular}{rccc}
    \toprule
    \multicolumn{1}{c}{\textbf{Verifiers}} & Aesthetic & CLIPScore & ImageReward \\
    \midrule
    \multicolumn{1}{c}{-}      & 5.79 &  0.71	&	0.97\\
    \midrule
    CLIP-SSL + 4 search iters & 5.76 & 0.71 & 0.99 \\
    + 16 search iters  & 5.72 & 0.71 & 1.04\\
    \midrule
    DINO-SSL + 4 search iters & 5.79 & 0.71 & 0.99 \\
    + 16 search iters & 5.78 & 0.70 & 1.03\\
    \midrule 
    SigLIP-SSL + 4 search iters & 5.79 & 0.70 & 1.02 \\
    + 16 search iters & 5.75 & 0.70 & 1.02\\
    \bottomrule
    \end{tabular}
    }
    \caption{\footnotesize{\textbf{\textit{Performance of self-supervised verifiers on DrawBench.}} All numbers are from FLUX.1-dev with random search. The first row is the reference performance without search.}}
    \label{tab:ssl-t2i}
    \vspace{-0.5cm}
\end{table}

\begin{figure}[h!]
    \centering
    \fbox{
        \begin{minipage}{1.0\textwidth}
            \ttfamily
            "You are a multimodal large-language model tasked with evaluating images generated by a text-to-image model. Your goal is to assess each generated image based on specific aspects and provide a detailed critique, along with a scoring system. The final output should be formatted as a JSON object containing individual scores for each aspect and an overall score. Below is a comprehensive guide to follow in your evaluation process:

            \textbf{1. Key Evaluation Aspects and Scoring Criteria:}
            
            For each aspect, provide a score from 0 to 10, where 0 represents poor performance and 10 represents excellent performance. For each score, include a short explanation or justification (1-2 sentences) explaining why that score was given. The aspects to evaluate are as follows:

            \textbf{a) Accuracy to Prompt}

            Assess how well the image matches the description given in the prompt. Consider whether all requested elements are present and if the scene, objects, and setting align accurately with the text.
            Score: 0 (no alignment) to 10 (perfect match to prompt).

            \textbf{b) Creativity and Originality}

            Evaluate the uniqueness and creativity of the generated image. Does the model present an imaginative or aesthetically engaging interpretation of the prompt? Is there any evidence of creativity beyond a literal interpretation?
            Score: 0 (lacks creativity) to 10 (highly creative and original).

            \textbf{c) Visual Quality and Realism}
            
            Assess the overall visual quality, including resolution, detail, and realism. Look for coherence in lighting, shading, and perspective. Even if the image is stylized or abstract, judge whether the visual elements are well-rendered and visually appealing.
            Score: 0 (poor quality) to 10 (high-quality and realistic).

            \textbf{d) Consistency and Cohesion}

            Check for internal consistency within the image. Are all elements cohesive and aligned with the prompt? For instance, does the perspective make sense, and do objects fit naturally within the scene without visual anomalies?
            Score: 0 (inconsistent) to 10 (fully cohesive and consistent).

            \textbf{e) Emotional or Thematic Resonance}

            Evaluate how well the image evokes the intended emotional or thematic tone of the prompt. For example, if the prompt is meant to be serene, does the image convey calmness? If it’s adventurous, does it evoke excitement?
            Score: 0 (no resonance) to 10 (strong resonance with the prompt's theme).
            
            \textbf{2. Overall Score}

            After scoring each aspect individually, provide an overall score, representing the model's general performance on this image. This should be a weighted average based on the importance of each aspect to the prompt or an average of all aspects."
        \end{minipage}
    }
    \caption{\footnotesize{\textbf{\textit{The detailed prompt for evaluation with the LLM Grader.}}}}
    \label{fig:gemini-prompt}
\end{figure}

\clearpage

\section{More Visualizations on Scaling Behavior}

\subsection{SiT-XL}

The images presented in this section are sampled from the pre-trained SiT-XL in $256$ resolution, using 2$^\text{nd}$ order Heun sampler and guidance scale of $4.0$. Each row of images is structured as follows: \begin{itemize}
    \item \textbf{Left three}: sampled with increasing steps: 10, 20, 250. 
    \item \textbf{Right three}: sampled with Zero-Order Search and the DINO classification verifier. We set $N = 2$ and $\lambda = 0.95$ for Zero-Order Search, and the equivalent NFEs invested are 450, 1850, 6650.
\end{itemize}

\begin{figure}[h]
    \captionsetup{singlelinecheck=false} 
    \centering
    \includegraphics[width=\linewidth]{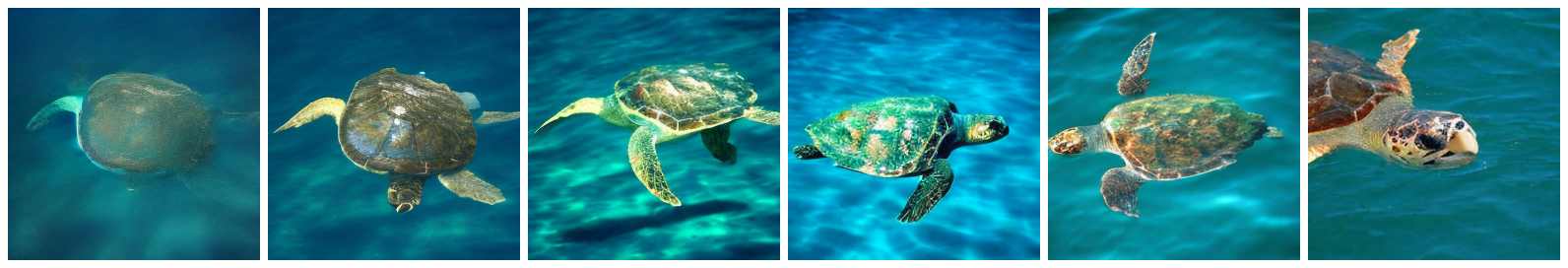}
    \caption{\textbf{``loggerhead turtle'' (33)}}
\end{figure}

\begin{figure}[h]
    \captionsetup{singlelinecheck=false} 
    \centering
    \includegraphics[width=\linewidth]{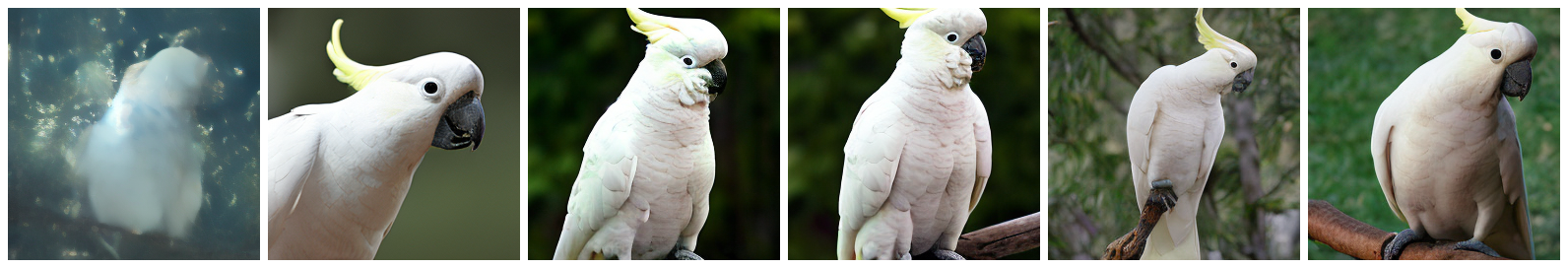}
    \caption{\textbf{``Sulphur-crested cockatoo'' (89)}}
\end{figure}

\begin{figure}[h]
    \captionsetup{singlelinecheck=false} 
    \centering
    \includegraphics[width=\linewidth]{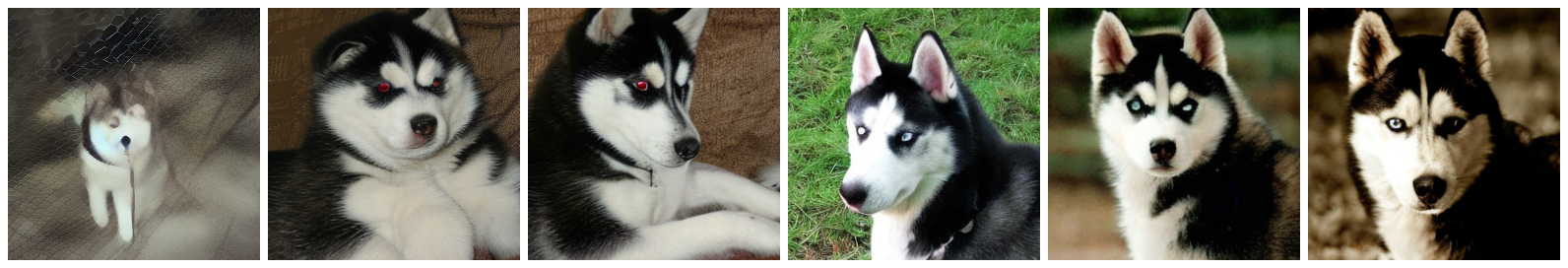}
    \caption{\textbf{``Siberian husky'' (250)}}
\end{figure}

\begin{figure}[h]
    \captionsetup{singlelinecheck=false} 
    \centering
    \includegraphics[width=\linewidth]{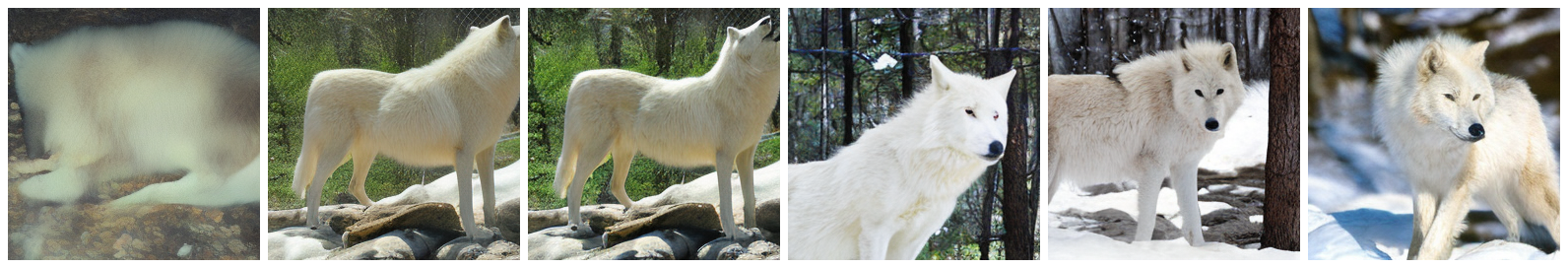}
    \caption{\textbf{``Arctic wolf'' (270)}}
\end{figure}

\begin{figure}[h]
    \captionsetup{singlelinecheck=false} 
    \centering
    \includegraphics[width=\linewidth]{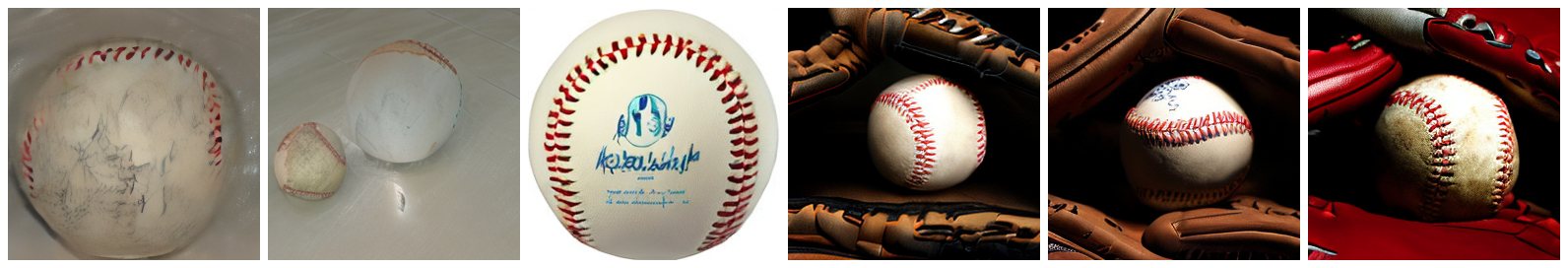}
    \caption{\textbf{``baseball'' (429)}}
\end{figure}

\begin{figure}[h]
    \captionsetup{singlelinecheck=false} 
    \centering
    \includegraphics[width=\linewidth]{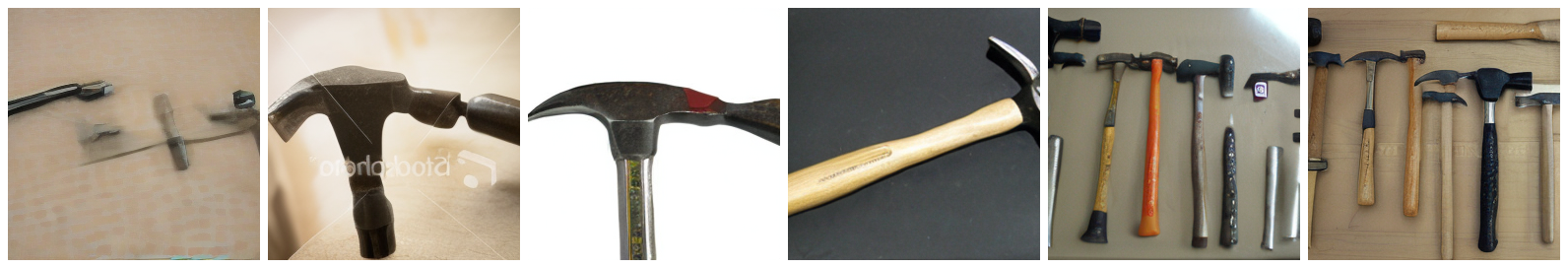}
    \caption{\textbf{``hammer'' (587)}}
\end{figure}

\begin{figure}[h]
    \captionsetup{singlelinecheck=false} 
    \centering
    \includegraphics[width=\linewidth]{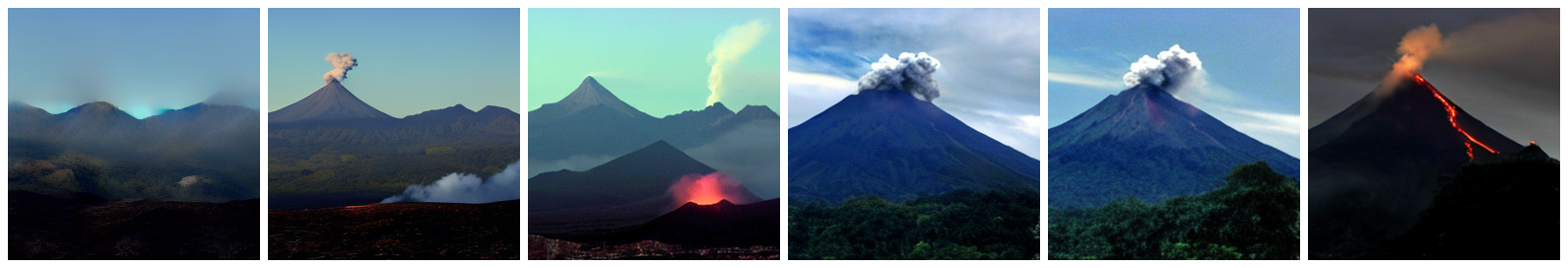}
    \caption{\textbf{``volcano'' (980)}}
\end{figure}

\clearpage

\subsection{FLUX.1-dev}

The images presented in this section are sampled from the pre-trained FLUX.1-dev in $1024$ resolution, using Euler sampler and guidance scale of $3.5$. Each row of images is structured as follows: \begin{itemize}
    \item \textbf{Left three}: sampled with increasing steps: 4, 16, 30. 
    \item \textbf{Right three}: sampled with Zero-Order Search and the Verifier Ensemble. We set $N = 2$ and $\lambda = 0.95$ for Zero-Order Search, and the equivalent NFEs invested are 120, 960, 2880.
\end{itemize}

\begin{figure}[h]
    \includegraphics[width=\linewidth]{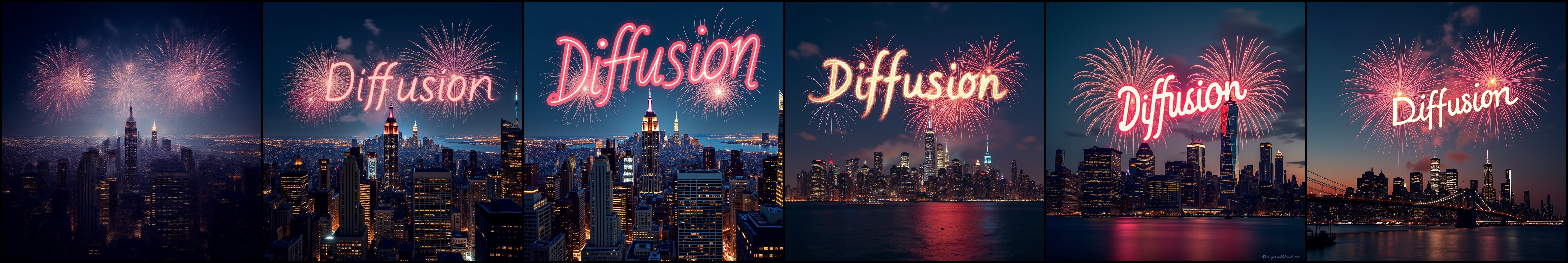}
    \caption{\textbf{``New York Skyline with `Diffusion' written with fireworks on the sky.''}}
\end{figure}

\begin{figure}[h]
    \captionsetup{singlelinecheck=false} 
    \centering
    \includegraphics[width=\linewidth]{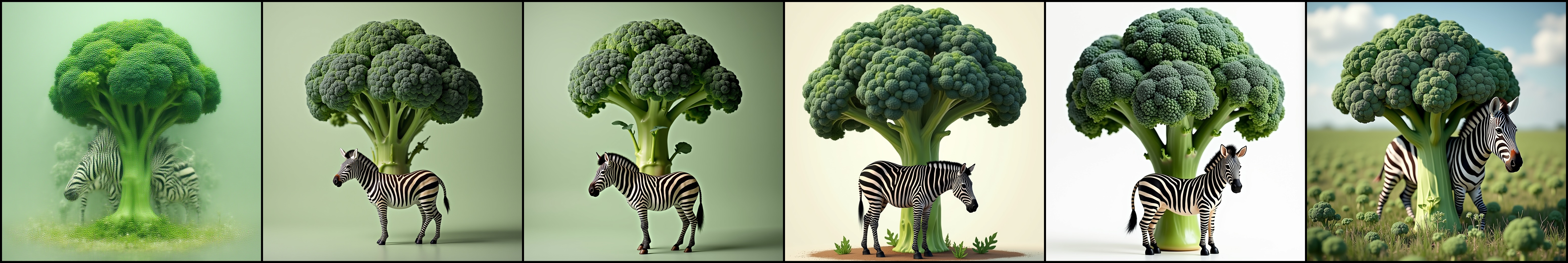}
    \caption{\textbf{``A zebra underneath a broccoli.''}}
\end{figure}

\begin{figure}[h]
    \captionsetup{singlelinecheck=false} 
    \centering
    \includegraphics[width=\linewidth]{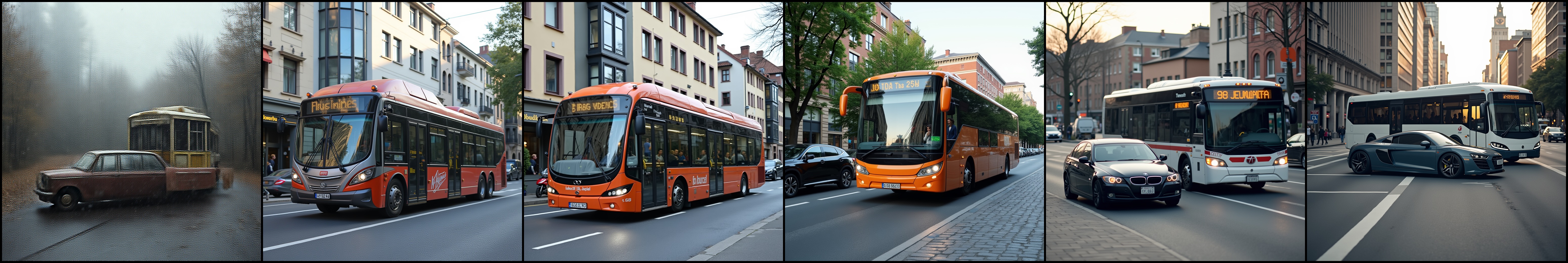}
    \caption{\textbf{``A car on the left of a bus.''}}
\end{figure}

\begin{figure}[h]
    \captionsetup{singlelinecheck=false} 
    \centering
    \includegraphics[width=\linewidth]{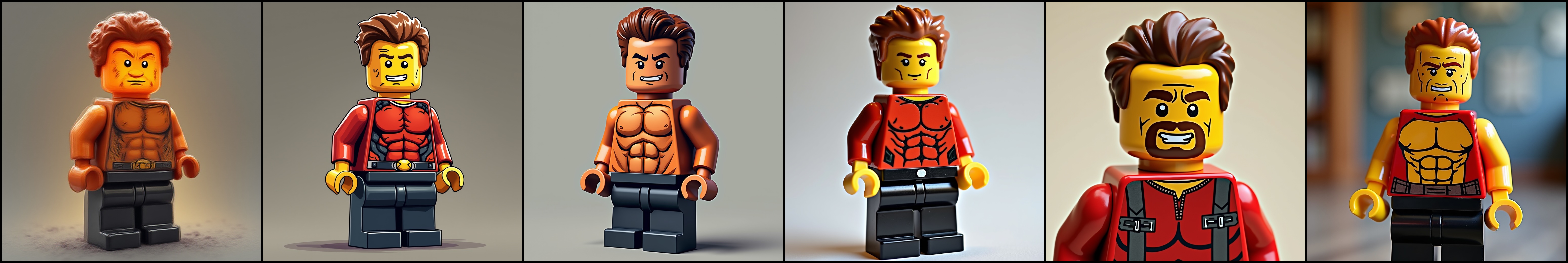}
    \caption{\textbf{``Lego Arnold Schwarzenegger.''}}
\end{figure}

\begin{figure}[h]
    \captionsetup{singlelinecheck=false} 
    \centering
    \includegraphics[width=\linewidth]{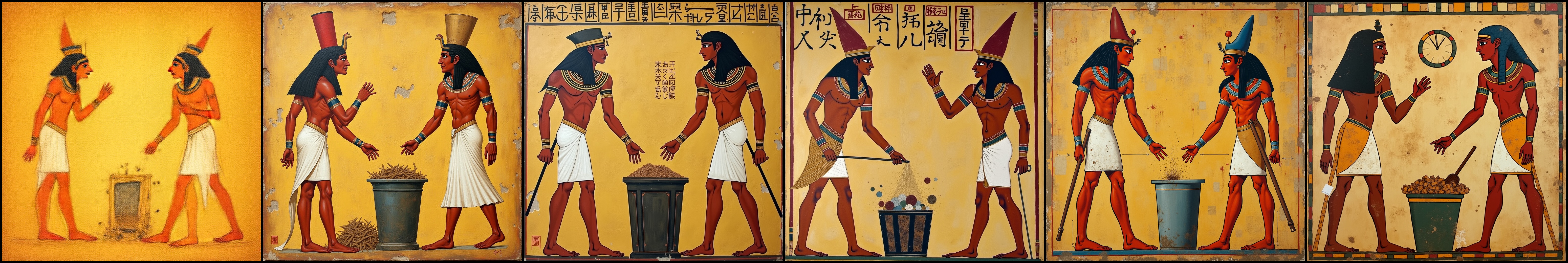}
    \caption{\textbf{``An ancient Egyptian painting depicting an argument over whose turn it is to take out the trash.''}}
\end{figure}

\begin{figure}[h]
    \captionsetup{singlelinecheck=false} 
    \centering
    \includegraphics[width=\linewidth]{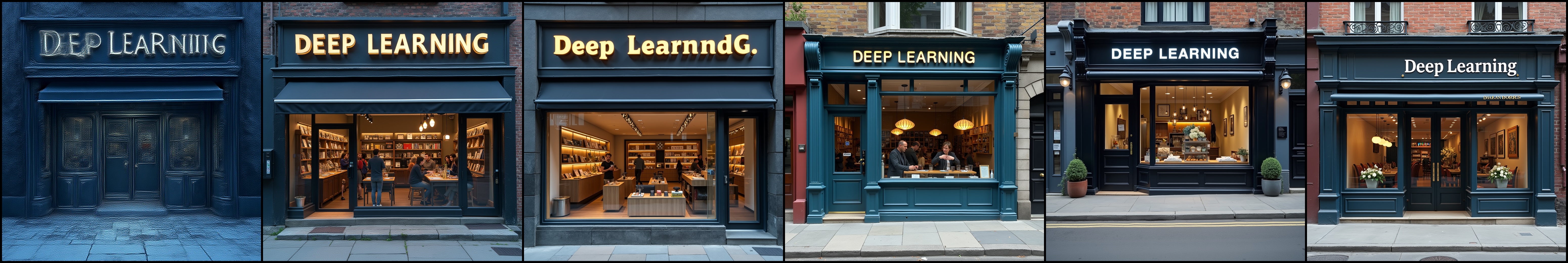}
    \caption{\textbf{``A storefront with `Deep Learning' written on it.''}}
\end{figure}

\begin{figure}[h]
    \captionsetup{singlelinecheck=false} 
    \centering
    \includegraphics[width=\linewidth]{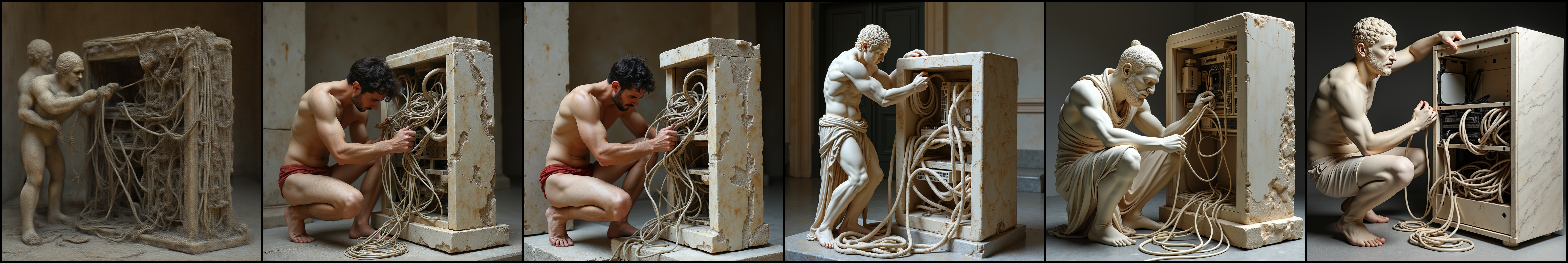}
    \caption{\textbf{``An IT-guy trying to fix hardware of a PC tower is being tangled by the PC cables like Laokoon. Marble, copy after Hellenistic original from ca. 200 BC. Found in the Baths of Trajan, 1506.''}}
\end{figure}

\clearpage

\subsection{PixArt-$\Sigma$}

The images presented in this section are sampled from the pre-trained PixArt-$\Sigma$ in $1024$ resolution, using DDIM sampler and guidance scale of $4.5$. Each row of images is structured as follows: \begin{itemize}
    \item \textbf{Left three}: sampled with increasing steps: 4, 8, 28. 
    \item \textbf{Right three}: sampled with Zero-Order Search and the Verifier Ensemble. We set $N = 2$ and $\lambda = 0.95$ for Zero-Order Search, and the equivalent NFEs invested are 112, 896, 2688.
\end{itemize}

\begin{figure}[h]
    \captionsetup{singlelinecheck=false} 
    \centering
    \includegraphics[width=\linewidth]{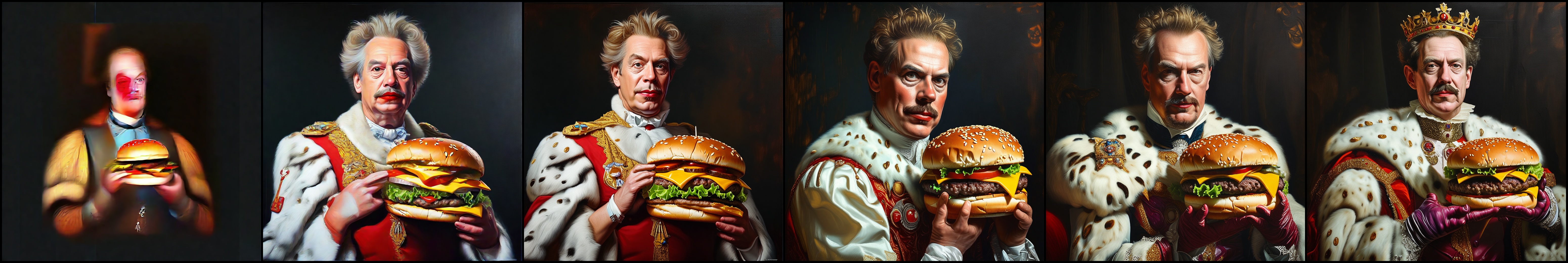}
    \caption{\textbf{``An oil painting portrait of the regal Burger King posing with a Whopper.''}}
\end{figure}

\begin{figure}[h]
    \captionsetup{singlelinecheck=false} 
    \centering
    \includegraphics[width=\linewidth]{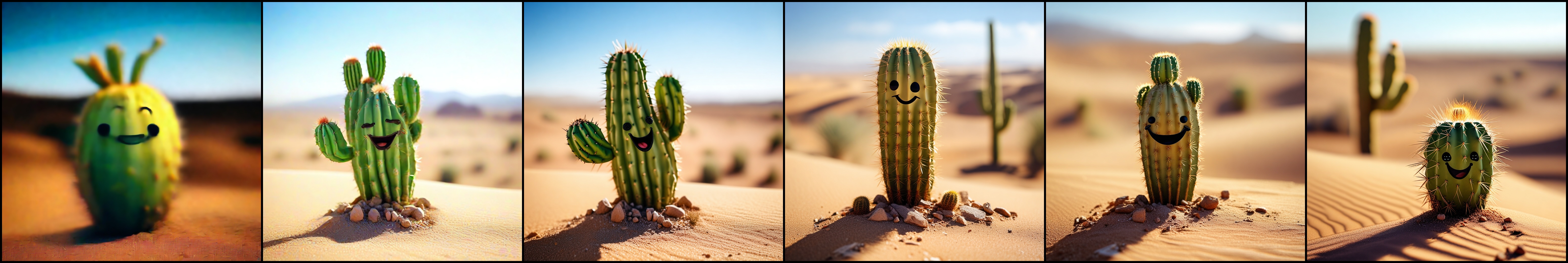}
    \caption{\textbf{``A small cactus with a happy face in the Sahara desert.''}}
\end{figure}

\begin{figure}[h]
    \captionsetup{singlelinecheck=false} 
    \centering
    \includegraphics[width=\linewidth]{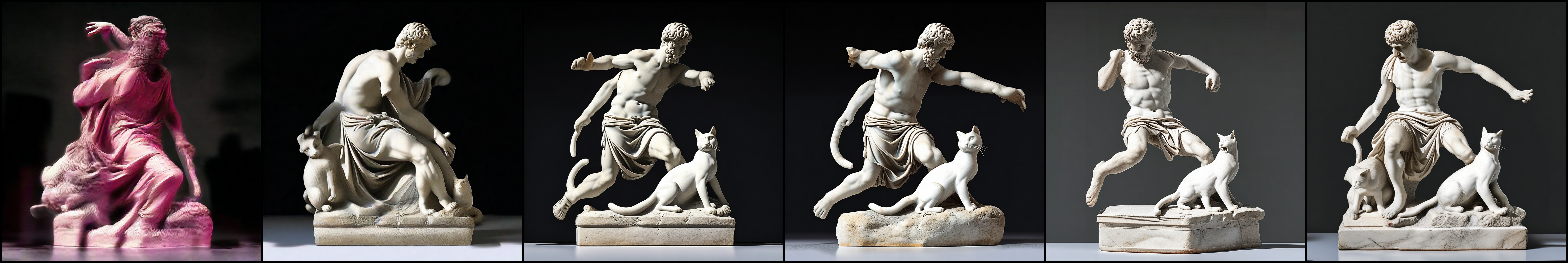}
    \caption{\textbf{``Greek statue of a man tripping over a cat.''}}
\end{figure}

\end{document}